\documentclass[10pt,journal,compsoc]{IEEEtran}
\usepackage[pdftex]{graphicx}
\usepackage{amsmath}
\usepackage{multirow}
\usepackage{multicol}
\usepackage{url}
\usepackage{enumitem}
\usepackage{booktabs}

\ifCLASSOPTIONcompsoc
  \usepackage[nocompress]{cite}
\else
  \usepackage{cite}
\fi
\ifCLASSINFOpdf
\else
\fi

\usepackage{amsfonts}
\usepackage{amssymb}
\usepackage{xcolor}

\begin{document}
%
\title{A Systematic Survey on Deep Generative Models for Graph Generation}
%
%
%
%

\author{Xiaojie Guo, Liang Zhao
\IEEEcompsocitemizethanks{\IEEEcompsocthanksitem Xiaojie Guo is with the IBM T. J. Watson Research Center, Yorktown Height, NY. 
E-mail:xguo7@gmu.edu\protect\\
\IEEEcompsocthanksitem Liang Zhao is with the Department
of Computer Science, Emory University, Atlanta,
GA. E-mail:liang.zhao@emory.edu.
Corresponding Author.\protect\\

}
}

\IEEEtitleabstractindextext{%
\begin{abstract}
Graphs are important data representations for describing objects and their relationships, which appear in a wide diversity of real-world scenarios. As one of a critical problem in this area, graph generation considers learning the distributions of given graphs and generating more novel graphs. Owing to their wide range of applications, generative models for graphs, which have a rich history, however, are traditionally hand-crafted and only capable of modeling a few statistical properties of graphs. Recent advances in deep generative models for graph generation is an important step towards improving the fidelity of generated graphs and paves the way for new kinds of applications. This article provides an extensive overview of the literature in the field of deep generative models for graph generation. Firstly, the formal definition of deep generative models for the graph generation and the preliminary knowledge are
provided. Secondly, taxonomies of deep generative models for both unconditional and conditional graph generation are proposed respectively; the existing works of each are compared and analyzed. After that, an overview of the evaluation metrics in this specific domain is provided. Finally, the applications that deep graph generation enables are summarized and five promising future research directions are highlighted.
\end{abstract}
}

\maketitle

\IEEEdisplaynontitleabstractindextext

%
\IEEEpeerreviewmaketitle

\IEEEraisesectionheading{\section{Introduction}\label{sec:introduction}}
\IEEEPARstart{G}{raphs} are ubiquitous in the real world, representing objects and their relationships such as social networks, citation networks, biology networks, traffic networks, etc. Graphs are also known to have complicated structures that contain rich underlying values ~\cite{barabasi2016network}. Tremendous efforts have been made in this area, resulting in a rich literature of related papers and methods to deal with various kinds of graph problems. These works can be categorized into two types: 1) predicting and analyzing patterns on given graphs. 2) learning the distributions of given graphs and generating more novel graphs. The first type covers many research areas including node classification, graph classification, and link prediction. Over the past few decades, a significant amount of work has been done in this domain. In contrast to the first type, the second type is related to graph generation problem, which is the focus of this paper.


Graph generation includes the process of modeling and generating real-world graphs, and it has applications in
several domains, such as understanding interaction dynamics in
social networks~\cite{grover2019graphite,wang2018graphgan,tran2019deepnc}, anomaly detection~\cite{ranu2009graphsig}, protein structure modeling~\cite{guo2020generating,du2022interpretable}, source code generation and translation~\cite{brockschmidt2019generative,dai2018syntax}, and semantic parsing~\cite{zhang2019amr}. Owing to its many applications, the development of generative models for graphs has a rich history, resulting in famous models such as random graphs, small-world models, stochastic block models, and Bayesian network models, which generate graphs based on apriori structural assumptions~\cite{newman2018networks}. These graph generation models~\cite{albert2002statistical,leskovec2010kronecker,robins2007recent} are engineered towards modeling a pre-selected family of graphs, such as random graphs~\cite{erdos1959random}, small-world networks~\cite{watts1998collective}, and scale-free graphs~\cite{albert2002statistical}. However, due to their simplicity and hand-crafted nature, these random graph models generally have limited capacity to model complex dependencies and are only capable of modeling a few statistical properties
of graphs. Such methods usually fit well towards the properties that the predefined principles are tailored for, but usually cannot do well for the others. For example, a contact network models can fit flu epidemics but not dynamic functional connectivity. However, in many domains, the network properties and generation principles are largely unknown, such as those for explaining the mechanisms of mental diseases in brain network, cyber-attacks, and malware propagations. For the other example, Erdos–Rényi graphs do not have the heavy-tailed degree distribution that is typical of many real-world networks. In addition, the utilization of the apriori assumption limits these traditional techniques from exploring more applications in larger scale of domains, where the apriori knowledge of graphs are always not available. 

Considering the limitations of the traditional graph generation techniques, a key open challenge is developing methods that can directly learn generative models from an observed set of graphs, which is an important step towards improving the fidelity of generated graphs. It paves the way for new kinds of applications, such as novel drug discovery~\cite{popova2019molecularrnn,you2018graphrnn}, and protein structure modeling~\cite{bacciu2019graph,anand2018generative,fan2019labeled}. Recent advances in deep generative models, such as variational autoencoders (VAE)~\cite{kingma2014auto} and generative adversarial networks (GAN)~\cite{goodfellow2014generative}, have supported a number of deep learning models for generating graphs have been proposed, which formalized the promising area of \emph{Deep Generative Models for Graph Generation}, which is the focus of this survey.

\vspace{-0.15cm}
\subsection{Formal Problem Definition}
A graph is defined as $G(\mathcal{V},\mathcal{E},F,E)$, where $\mathcal{V}$ is the set of $N$ nodes, and $\mathcal{E}\subseteq \mathcal{V} \times \mathcal{V}$ is the set of $M$ edges. $e_{i,j}\in\mathcal{E}$ is an edge connecting nodes $v_i, v_j\in\mathcal V$. The graph can be conveniently described in the form of matrix or tensor using its (weighed) adjacency matrix $A$. If the graph is node-attributed or edge-attributed, there are node attribute matrix $F\in \mathbb R^{ N\times D}$ assigning attributes to each node or edge attribute tensor $E\in \mathbb R^{ N\times N\times K}$  assigning attributes to each edge $e_{i,j}$. $K$ is the dimension of the edge attributes, and $D$ is the dimension of the node attributes. 

Given a set of observed graphs $\mathbb{G} = \{G_1,...G_s\}$ sampled from the data distribution $p(G)$, where each graph $G_i$ may have different numbers of nodes and edges, the goal of learning generative models for graphs is to learn the distribution of the observed set of graphs.
By sampling a graph $G\sim p_{model}(G)$, new graphs can hence be achieved, which is known as deep graph generation, the short form of deep generative models for graph generation. Sometimes, the generation process can be conditioned on additional information $y$, such that $G\sim p_{model}(G|y)$, in order to provide extra control over the graph generation results. The generation process with such conditions is called conditional deep graph generation.

\vspace{-0.15cm}
\subsection{Challenges}
The development of deep generative models for graphs poses unique challenges, which are mainly listed below.


\textbf{Non-unique Representations}.
In the general setting, a graph with $n$ nodes can be represented by up to $n!$ equivalent adjacency matrices,
each corresponding to a different, arbitrary node ordering. 
Given that a graph can have multiple representations, it is difficult for the models to calculate the distance between the generated graphs and ground-truth graphs white training. Thus it may require us to design either a pre-defined node ordering or a node permutation invariant reconstruction objective function.

\textbf{Complex Dependencies}. The nodes and edges of a graph have complex dependencies and relationships. For example, two nodes are more likely to be connected if they share common neighbors. Therefore, the generation of each node or edge cannot be modeled as an independent event. One way to formalize the graph generation is to make auto-regressive decisions, which naturally accommodate complex dependencies inside the graphs through sequential formalization of graphs. Towards this challenge, in this survey, existing works are described and compared regarding to what kinds of dependencies (e.g., dependencies among nodes, among edges or between node and edges) they can capture.


\textbf{Large Output Spaces}.
To generate a graph with $n$ nodes the generative model may have to output $n^2$ values to specify its structure, which makes it expensive, especially for large-scale graph. However, it is common to find graphs containing millions of graphs in real-world, such as citation and social networks. Consequently, it is important for generative models to scale to large-scale graphs for realistic graph generation and to accommodate such complexity in the output space. The scalability of the existing works is a critical issue in comparing and evaluating the different categories of graph generative models in this survey, as discussed in Section~\ref{sec:2.1.5} and Section~\ref{sec:2.3.2}.

\textbf{Discrete Objects by Nature}. 
The standard machine learning techniques, which were developed primarily for continuous data, do not work off-the-shelf, but usually need adjustments. A prominent example is the back-propagation algorithm, which is not directly applicable to graphs, since it works only for continuously differentiable objective functions. To this end, it is usual to project graphs (or their constituents) into a continuous space and represent them as vectors/matrix. However, reconstructing the generated graphs from the continuous representations is a challenge. Reconstructing the desecrate graph objects (i.e., nodes and edges) from continuous spaces results into different graph decoder process, such as sequentially generating the nodes of the graphs or generating the adjacent matrix of graphs in one-shot. This challenge motivates the major criteria in the taxonomy of the existing methods in this survey.


\textbf{Evaluation for Implicit Properties}
Evaluating the generated graphs is a very critical but challenging issue, due to the unique properties of graphs which with complex and high-dimensional structure and implicit features. Existing methods use different evaluation metrics. For example, some works~\cite{you2018graphrnn,sun2019graph,guo2022deep} compute the distance of the graph statistic distribution of the graphs in the test set and graphs that are generated, while other works~\cite{liuauto,fan2019labeled} indirectly use some classifier-based metrics to judge whether the generated graphs are of the same distribution as the training graphs. 
It is important to systematically review all the existing metrics and choose the approximate ones based on their strengths and limitations according to the application requirements. Towards this challenge, we summarized a unified evaluation framework for graph generation in Section~\ref{sec:five}, including popular evaluation metrics for both unconditional and conditional graph generation.

\textbf{Various Validity Requirements}.
Modeling and understanding graph generation via deep learning involve a wide variety of important applications, including molecule designing~\cite{popova2019molecularrnn,jin2018junction}, protein structure modeling~\cite{anand2018generative}, AMR parsing~\cite{lyu2018amr,zhang2019amr}, et al. These inter-discipline problems have their unique requirements for the validity of the generated graphs. For example, the generated molecule graphs need to have valency validity, while the semantic parsing in NLP requires Part-of-Speech (POS)-related constraint. Thus, addressing the validity requirements for different applications is crucial in enabling wider applications of deep graph generation. In this paper, we elaborate the way how the existing works improve the validity of the generated graph when introducing the rule-based generation models in Section~\ref{sec:rule-based}. In addition, the real-world applications of solving validity issues are elaborated in Section\ref{sec:six}.

\textbf{Black-box with Low Reliability}.
Compared with the traditional graph generation area, deep learning based graph modeling methods are like black-boxes which bear the weaknesses of low interpretability and reliability. 
Improving the interpretability of the deep graph generative models could be a vital issue in unpacking the black-box of the generation process and paving the way for wider application domains, which are of high sensitivity and require strong reliability, such as smart health and automatic driving. In addition, semantic explanation of the latent representations can further enhance the scientific exploration of the associated application domains. Interpretability and reliability are important aspects when comparing and evaluating the different graph generation methods in this survey, as discussed in Section~\ref{sec:3.1.3}, which compares the different conditional graph generation categories. 

\begin{figure*}[tb]
 \centering
 \includegraphics[width=0.9\textwidth]{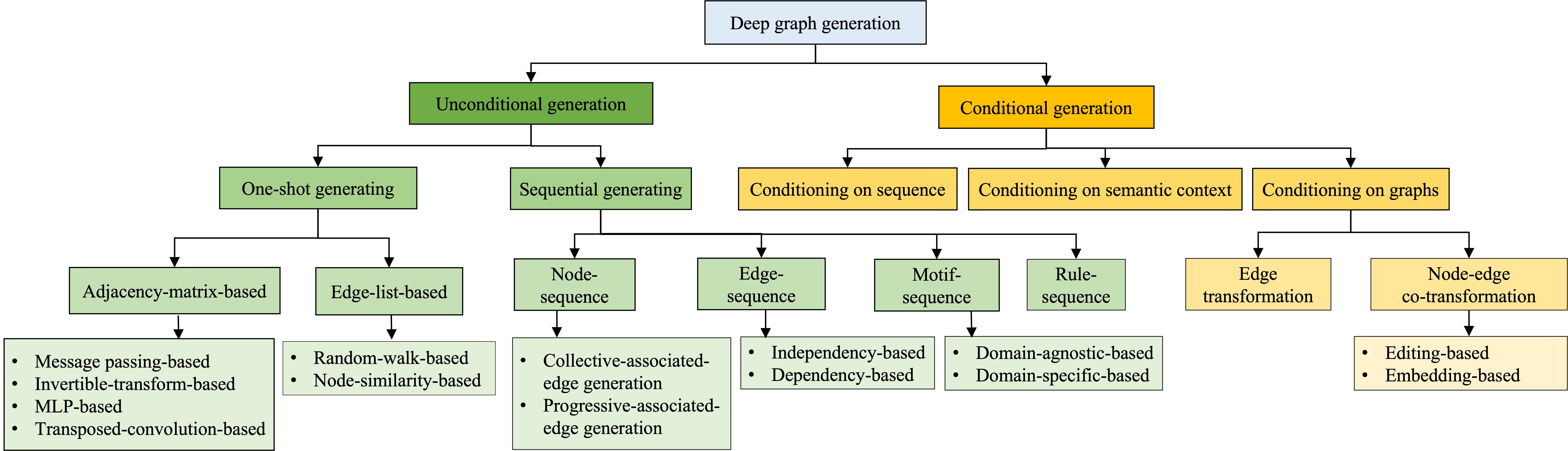}
 \caption{Classification of deep generative models for graph generation problems}
  \label{fig:tree}
\end{figure*}

\vspace{-0.15cm}
\subsection{Our Contributions}
Various advanced works on deep graph generation have been conducted, ranging from the one-shot graph generation to sequential graph generation process, accommodating various deep generative learning strategies. These methods aim to solve one or several of the above challenges by works from different fields, including machine learning, bio-informatics, artificial intelligence, human health and social-network mining. However, the methods developed by different research fields
tend to use different vocabularies and solve problems from
different angles. Also, standard and comprehensive evaluation procedures to validate the developed deep generative models for graphs are lacking. 

To this end, this paper provides a systematic review of deep generative models for graph generation. 
The goal is to help interdisciplinary researchers choose appropriate techniques to solve problems in their applications domains, and more importantly, to help
graph generation researchers understand the basic principles as well as identify open research opportunities in deep graph generation domain. As far as we  know, this is the first comprehensive survey on deep generative models for graph generation. Below, we summarize the major contributions of this survey:
\begin{itemize}[leftmargin=*]
    \item We propose a taxonomy of deep generative models for graph generation categorized by problem settings and methodologies. The drawbacks, advantages, and relations among different subcategories have been introduced.
    \item We provide a detailed description, analysis, and comparison of deep generative models for graph generation as well as the base deep generative models.
    \item We summarize and categorize the existing evaluation procedures and metrics, the benchmark datasets and the corresponding results of deep generative models for graph generation tasks.
    \item We introduce existing application domains of deep generative models for graphs as well as the potential benefits and opportunities they bring into these applications.
    \item We suggest several open problems and promising future research directions in the field of deep generative models for graph generation.
\end{itemize}

\subsection{Relationship with Deep Generative Models}
Deep generative models form the backbone of the base learning methods of all the existing deep generative models for graph generation. Specifically, deep generative models offer a very efficient way to analyze and understand unlabeled data. The idea behind generative models is to capture the inner probabilistic distribution that generates a class of data to generate similar data~\cite{oussidi2018deep}. Emerging approaches such as generative adversarial networks (GANs)~\cite{goodfellow2014generative}, variational auto-encoders (VAEs)~\cite{kingma2014auto}, generative recursive neural network (generative RNN)~\cite{sutskever2011generating} (e.g., pixelRNNs, RNN language models), flow-based learning~\cite{papamakarios2017masked}, and many of their variants have led to impressive results in myriads of applications. We provide a review of five popular and classic deep generative models in Appendix A.

\vspace{-0.3cm}
\subsection{Relationship with Existing Surveys}
There are three types of existing surveys that are relevant to our work. The first type mainly centers around the traditional graph generation by classic graph theory and network science~\cite{bonifati2020graph}, which does not focus on the most recent advancement in deep generative neural networks in AI. The second type is about representation learning on graphs~\cite{goyal2018graph, wu2020comprehensive, zhang2020deep}, which focuses on learning graph embedding given existing graphs. Few works include a handful of deep generative models that could be used for representation learning tasks. The third type is specific to particular applications such as molecule design by deep learning, instead of for this generic technical domain. 

As yet, there have been very few systematic surveys on deep generative models for graph generation, with just two recent contemporaneous papers~\cite{faez2021deep,zhu2022survey}. Both of these categorize graph generation mainly in terms of the general backbone learning models utilized (i.e., auto-regressive, auto-encoder-based, RL-based, adversarial, and flow-based), We have instead opted to review this research field from more comprehensive and graph-specific perspectives, including task formulation, graph generating techniques, evaluations, applications and datasets. This yields a number of advantages compared to the existing ones, namely: (1) \textbf{Two main problems are covered}: This survey comprehensively summarizes the techniques used for both unconditional and conditional generation problems; (2) \textbf{Categorization from graph-specific perspectives}: This survey categorizes the existing graph generation models (e.g., sequential-generating and one-shot generation) utilizing graph-specific perspectives, instead of the all-in purpose generative models developed and applied for all kinds of data generation; (3) \textbf{Reviews of evaluation methods}: This survey provides a comprehensive overview of the existing evaluation procedures and metrics for graph generation tasks;
(4) \textbf{More applications}: This survey provides a comprehensive summary for a diverse range of the applications, including domains like biology, NLP and program analysis; and
(5) \textbf{Performance comparisons}: This survey compares the performance of existing state-of-the-arts methods on both synthetic and real-world datasets, reaching several insightful conclusions.

\vspace{-0.2cm}
\subsection{Outline of the Survey}
The remaining part of the survey is organized as follows. 
In Sections \ref{sec:three} and \ref{sec:four}, we provide the taxonomy of
deep graph generation, and the taxonomy structure is illustrated
in Fig.~\ref{fig:tree}. Section~\ref{sec:three} compares related works of unconditional deep graph generation problem and summarizes the challenges faced in each. In Section~\ref{sec:four}, we categorize the conditional deep graph generation in terms of three sub-problem settings. The challenges behind each problem are summarized, and a detailed analysis of different techniques is provided. Lastly, we summarize and categorize the evaluation metrics in Section~\ref{sec:five}.
Then we present the applications that deep graph generation enables in Section~\ref{sec:six}. At last, we discuss five potential
future research directions and conclude this
survey in Sections~\ref{sec:seven} and ~\ref{sec:eight}.
Due to the space limit, We also summarize the benchmark dataset and performance evaluation of existing works in Appendix B. 

\section{Unconditional Deep Generative Models for Graph Generation}
\label{sec:three}

The goal of unconditional deep graph generation is to learn the distribution $p_{\mathrm{model}}(G)$ based on a set of observed realistic graphs being sampled from the real distribution $p(G)$ by deep generative models. 
Based on the style of the generation process, we can categorize the methods into two main branches: (1) \textbf{Sequential generating}: this generates the nodes and edges in a sequential way, one after another, (2) \textbf{One-shot generating}: this refers to building a probabilistic graph model based on the matrix representation that can generate all nodes and edges in one shot. These two ways of generating graphs have their limitations and merits.
Sequential generating performs the local decisions made in the preceding one in an efficient way, but it has difficulty in preserving the long-term dependency. Thus, some global properties (e.g., scale-free property) of the graph are hard to include. Moreover, existing works on sequential generating are limited to a predefined ordering of the sequence, leaving open the role of permutation. 
One-shot generating methods have the capacity of modeling the global property of a graph by generating and refining the whole graph (i.e. nodes and edges) synchronously through several iterations, but most of them are hard to scale to large graphs since the time complexity is usually over $\textbf{O}(N^2)$ because of the needs of collectively modeling global relationship among nodes.
\small 
\begin{table*}[htb]
    \centering    
    \caption{Deep Generative-based Methods for Unconditional Graph Generation}\vspace{-0.35cm}
    \begin{tabular}{|l|l|l|l|}
    \toprule\hline
         Generating Style&\multicolumn{2}{|l|}{Techniques}&Reference \\\hline
         \multirow{4}{*}{Sequential Generating}&\multirow{2}{*}{Node-sequence-based}&Collective-associated-edge-generation&\cite{khodayardeep,d2019deep,zhang2019d,you2018graphrnn,su2019graph,popova2019molecularrnn, assouel2018defactor}\\\cline{3-4}
         ~&~&Progressive-associated-edge-generation&\cite{lim2019scaffold,li2018learning,liu2018constrained,kearnes2019decoding}\\\cline{2-4}
         ~&\multirow{2}{*}{Edge-sequence-based}&Independency-based&\cite{goyal2020graphgen}\\\cline{3-4} 
         ~&~&Dependency-based&\cite{bacciu2019graph,bacciu2020edge}\\\cline{2-4} 
         ~&\multirow{2}{*}{Motif-sequence-based}&Domain-agnostic-based&\cite{liao2019efficient}\\\cline{3-4} 
        ~&~&Domain-specific-based&\cite{jin2018junction,podda2020deep,gu2019explore}\\\cline{2-4} 
         ~&\multicolumn{2}{|l|}{Rule-sequence-based}&\cite{dai2018syntax,kusner2017grammar}\\\hline
         \multirow{5}{*}{One-shot Generating}&\multirow{3}{*}{Adjacency-matrix-based}&MLP-based&\cite{simonovsky2018graphvae,ma2018constrained,anand2018generative,fan2019labeled,polsterl2019likelihood,de2018molgan}\\\cline{3-4}
         ~&~&Message-Passing-based&\cite{bresson2019two,guarinodipol,flam2020graph,niu2020permutation}\\\cline{3-4}
         ~&~& Invertible-transform-based&\cite{honda2019graph,madhawa2019graphnvp}\\ \cline{3-4}
         ~&~& Transposed-convolution-based&\cite{guo2022deep,gao2018local}\\ \cline{2-4}
         ~&\multirow{2}{*}{Edge-list-based}&Random-walk-based&\cite{bojchevski2018netgan,gamage2020multi,zhang2019stggan,carida2019can}\\\cline{3-4}
          ~&~&Node-similarity-based&\cite{kipf2016variational,grover2019graphite,shi2020graphaf,zou2019encoding,liu2019graph,salha2019gravity}\\
          \hline\hline         
    \end{tabular}
    \label{tab:tox_unconditional}\vspace{-0.4cm}
\end{table*}\normalsize

\subsection{Sequential generating}
This type of methods treats graph generation as a sequential decision making process, wherein nodes and edges are generated one by one (or group by group), conditioning on the sub-graph already generated.
By modeling graph generation as a sequential process, these approaches naturally accommodate complex local dependencies between generated edges and nodes. A graph $G$ is represented by a sequence of components $S=\{s_1,...,s_N\}$, where each $s_i\in S$ can be regarded as a generation unit. The distribution of graphs $p(G)$ can then be formalized as the joint (conditional) probability of all the components in general. While generating graphs, different components will be generated sequentially, by conditioning on the already generated parts. 

One core issue is how to break down the graph generation to facilitate the sequential generation of its components, namely determining the formalization unit $s_i$ for sequentialization. The most straightforward approach is to formalize the graph as a sequence of nodes, which are the basic components of a graph, to support the \textit{node-sequence-based} generation. These methods essentially generates the graph by generating each node and its $O(N)$ associated edges in turn, and hence usually result in a total complexity of $O(N\cdot N)=O(N^2)$. Another approach is to consider a graph as set of edges, based on which a number of \textit{edge-sequence-based} generation methods have been proposed. These methods represent the graph as a sequence of edges and generate an edge, as well as its two ending nodes, per step, which leads to a total complexity of $O(|\mathcal{E}|\cdot 2)=O(|\mathcal{E}|)$. Edge-sequence-based methods are thus usually better at sparser graphs than node-sequence-based approaches. Although both these two types are successful at retaining pairwise node-relationships, they often fall short when it comes to capture higher-order relationships~\cite{gamage2020multi}. For example, gene regulatory networks, neuronal networks, and social networks all contain a large number of triangles; and molecular graphs contain functional groups. These all indicate the need to generalize the units of the sequential generation from nodes/edges to interesting sub-graph patterns, known as motifs. To this end, a number of \textit{motif-sequence-based} methods have been proposed that represent a graph utilizing a sequence of graph motifs so that a block of nodes and edges in a graph motif are generated simultaneously in each step, which usually boosts better efficiency. Although the above three types are all versatile in end-to-end graph generation, they fall short in ensuring generating ``valid'' graphs, namely graphs that enforce correct grammar and constraints, which are very common in fields like programming languages and molecules modeling. To solve this, several \textit{rule-sequence-based} methods have been proposed for domain specific applications, where a graph is constructed based on a predefined sequence of rules by incorporating appropriate domain expertise. A more detailed description of methods in each category is provided below.

\begin{figure}[htb]
    \centering
    \includegraphics[width=0.45\textwidth]{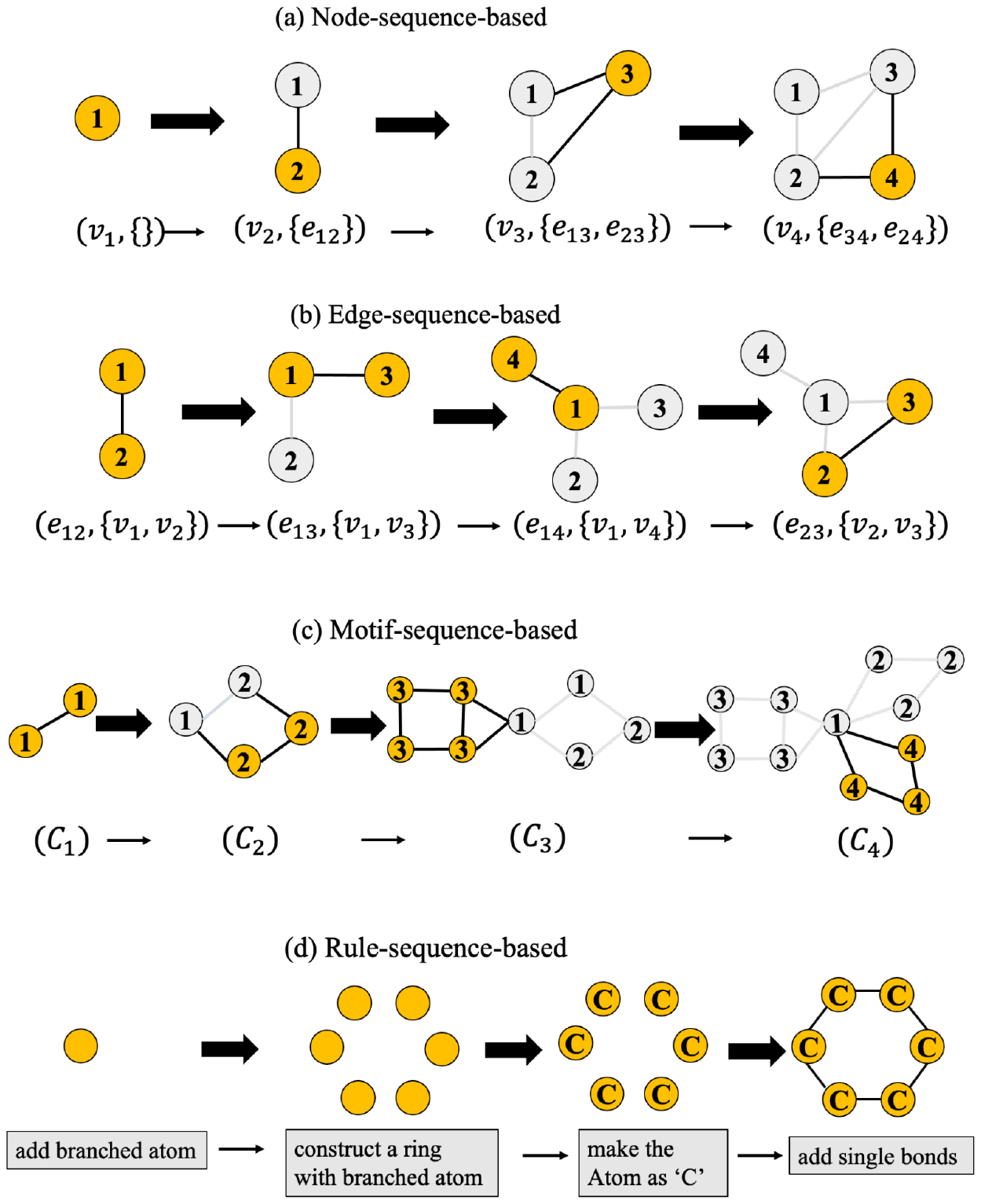}\vspace{-0.4cm}
    \caption{Four categories in graph sequential generating: the upper line of each sub-figure refers to the immediate graph that are generated per step; the bottom line of each sub-figure refers to the sequence consisting of unit $S_i$ that is generated per step.}\vspace{-0.4cm}
    \label{fig:sequence-generation}
\end{figure}

\subsubsection{Node-sequence-based}
\label{sec:node-sequence}

\textbf{General Framework}. Node-sequence-based methods essentially generate the graph by generating one node and its associated edges in each step, as shown in Fig.~\ref{fig:sequence-generation}(a). The graph is modeled by a sequence based on a predefined ordering $\pi$ on the nodes. Each unit $s_i$ in the sequence of components $S$ is represented as a tuple $s_i=(v^\pi_i,\{e_{i,j}\}_{j<i})$ (as shown at the bottom of Fig.~\ref{fig:sequence-generation}(a)), indicating that at each high-level step, the generator generates one node $v^\pi_i$ and all its associated edges set $\{e_{i,j}\}_{j<i}$\footnote{Here we omit the node and edge attribute symbol for clarity, but it is important to bear in mind that the generated node and edges can all have attributes (i.e. type, label).}. Specifically, in node-sequence-based generation, generating a unit $s_i$ involves two main steps. In the first step, a node is generated conditioning on the current generated graph $G_i$, which can be interpreted to learn $p(v^\pi_i|G_i)$. The second step is to generate the associated edges set $\{e_{i,j}\}_{j<i}$ for node $v_i^\pi$.

There are two options when it comes to generating the associated edges of each node: 1) collective associated-edge generation, where the predictions are conducted on all of the node pairs between $v_i^\pi$ and the other existing nodes in $G_i$ in a single shot to directly generate the associated edges set $\{e_{i,j}\}_{j<i}$; and 2) progressive-associated-edge generation, which generates the associated edges of node $v_i^\pi$ in sequence, with two actions per step: \emph{addEdge}, which determines the size of $\{e_{i,j}\}_{j<i}$, and \emph{selectNode}, which determines to which node the node $v_i^\pi$ will be connected if \emph{addEdge} is needed.\\


\noindent\textbf{Collective associated-edge generation}.
To conduct the predictions on node pairs between the newly generated node $v_i^\pi$ and all the other existing nodes, most of the works~\cite{khodayardeep,d2019deep,zhang2019d,you2018graphrnn,su2019graph,shi2020graphaf,popova2019molecularrnn} resort to predicting the adjacent vector $A^\pi_{i,\cdot}$, which covers all the potential edges from the newly added node $v_i$ to the other existing nodes. Thus, we can further represent each unit as $s_i=(v_i^\pi,A^\pi_{i,\cdot})$. 
And the sequence can be represented as $Seq(G,\pi)=\{(v_1^\pi,A_{1,\cdot}^\pi),...,(v_N^\pi,A_{N,\cdot}^\pi)\}$. The aim is to learn the distribution as:\vspace{-0.1cm}\small
\begin{equation}\vspace{-0.1cm}
    p(\mathcal{V}^\pi,A^\pi)=\prod\nolimits^{N}_{i=1}p(v^\pi_i|v^\pi_{<i},A^\pi_{<i,\cdot})p(A^\pi_{i,\cdot}|v^\pi_{\le i},A^\pi_{<i,\cdot}),\label{eq:12}
\end{equation} \normalsize
where $v^\pi_{<i}$ refers to the nodes generated before $v^\pi_i$ and $A^\pi_{<i,\cdot}$ refers to the adjacent vectors generated before $A^\pi_{i,\cdot}$. Such joint probability can be implemented by sequential-based architectures such as generative RNN models~\cite{you2018graphrnn,liuauto,popova2019molecularrnn,zhang2019d} and auto-regressive flow-based learning models~\cite{shi2020graphaf}. Here we introduce the RNN-based models as an example.

In the generative RNN-based models, the node distributions $p(v^\pi_i|v^\pi_{<i},A^\pi_{<i,\cdot})$ are typically assumed as a multivariate Bernoulli distribution that is parameterized by $\phi_i\in \mathbb{R}^{T}$, where $T$ refers to the number of node categories. The edge existence distribution $p(A^\pi_{i,\cdot}|v^\pi_{\le i},A^\pi_{<i,\cdot})$ can be assumed as the joint probability of several dependent Bernoulli distributions:\vspace{-0.2cm}\small
\begin{equation}\vspace{-0.1cm}
  p(A^\pi_{i,\cdot}|A^\pi_{<i,\cdot})=\prod\nolimits^{i-1}_{j=1}p(A^\pi_{i,j}|A^\pi_{i,<j},A^\pi_{<i,\cdot}),\label{eq:13}
\end{equation}\normalsize
where $p(A^\pi_{i,\cdot}|A^\pi_{<i,\cdot})$ is parameterized by $\theta_i\in \mathbb{R}^{i-1}$ and the distribution of $p(A^\pi_{i,j}|A^\pi_{i,<j},A^\pi_{<i,\cdot})$ is parameterized by each entry $\theta_{i,j}$ in $\theta_i$. 
The architecture for implementing Eq.~\eqref{eq:12} and~\eqref{eq:13} can be regarded as a hierarchical-RNN, where the outer RNN is used for generating the nodes and the inner RNN is used for generating each node's associated edges. After either a node or edge is generated, a graph-level hidden representation of the already generated sub-
graph is updated through a message passing neural network (MPNN)~\cite{gilmer2017neural}. 
Specifically, at each Step $i$, a parameter $\phi_i$ will be calculated through a multilayer perceptron
(MLP)-based function based on the current graph-level hidden representation. The parameter $\phi_i$ is used to parameterize the Bernoulli distribution of node existence, from which node $v^\pi_i$ is sampled. After that, the adjacent vector $A^\pi_{i,\cdot}$ is generated by sequentially generating each of its entry. \\

\noindent\textbf{Progressive associated-edge generation}.
The above-introduced collective associated-edge generation has a time complexity of $O(N^2)$ that is time-consuming especially for sparse graphs. A remedy is to progressively select the nodes to be connected with the current node $v_i^\pi$ from the existing nodes $v_{<i}^{\pi}$, until the desired number of nodes is selected, which is small for sparse graph. Specifically, for the current node $v_i^\pi$, we generate $\{e_{i,j}\}_{j<i}$ by applying two functions: 1) an \emph{addEdge} function to determine the size of the edge set $\{e_{i,j}\}_{j<i}$ of node $v_i^\pi$ and 2) a \emph{selectNode} function to select the nodes to be connected from the existing graph ~\cite{lim2019scaffold,li2018learning,liu2018constrained,kearnes2019decoding}. The complexity of progressive associated-edge generation method is $O(MN)$ where $M$ refers to the number of edges. 

Specifically, after generating a node $v_i^\pi$ in the first step, an \emph{addEdge} function is used to output a parameter as $f_{addEdge}(h_{v_i}^\pi)$, following a Bernoulli distribution indicating whether to add an edge to the node $v_i^\pi$. Here $h_{v_i}^\pi$ refers to the node-level hidden states of $v_i^\pi$ which is calculated through a node embedding function, e.g., MPNN~\cite{gilmer2017neural} based on the already-generated parts of the graph. If an edge is determined to be added, the next step is selecting the neighboring node $v_j^\pi$ from the existing nodes. To achieve this, we can compute a score $m^\pi_j$ (as Eq.~\eqref{eq:score}) for each existing node $v_j^\pi$ based on \emph{selectNode} function $f_{selectNode}$, which is then passed through a softmax function~\cite{bishop2006pattern} to be properly normalized into a distribution of nodes: \vspace{-0.3cm}

\small
\begin{align}
  &m_{i,j}^\pi=f_{selectNode}(h_{v_i}^\pi,h_{v_j}^\pi), \quad where j<i. \label{eq:score} \\  
  &p(e_{i,j}|v^\pi_{<i},\{e_{<i,j}\}_{j<i})=softmax(m_{i,j}^\pi).\label{eq:softmax}
\end{align}\normalsize

The MLP-based function $f_{selectNode}$ maps pairs of node-level hidden states $h_{v_i}^\pi$ and $h_{v_j}^\pi$ to a score $m_{i,j}^\pi$ for connecting node $v_j^\pi$ to the new node $v_i^\pi$. This can be extended to handle discrete edge attributes by making $m_{i,j}^\pi$ a vector of scores with the same size as the number of the edge attribute's categories, and taking the softmax over all categories of the edge attribute. Based on the aforementioned procedure, the two functions $f_{addEdge}$ and $f_{selectNode}$ are iteratively executed to generate the edges within the edge set $\{e_{<i,j}\}_{j<i}$ of node $v^\pi_i$ until the terminal signal from function $f_{addEdge}$ indicates that no more edges for node $v_i$ are yet to be added.

\subsubsection{Edge-sequence-based}

\textbf{General Framework}. Edge-sequence-based methods~\cite{goyal2020graphgen,bacciu2019graph,bacciu2020edge} consider a graph to be a sequence of edges and generate an edge, along with its two end nodes, in each step, as shown in Fig.~\ref{fig:sequence-generation}(b). It defines an ordering of the edges in the graph and also an ordering function $\alpha(\cdot)$ for indexing the nodes. The graph $G$ can then be modeled by a sequence of edges, with each unit in the sequence being a tuple represented by $s_i=(\alpha(u),\alpha(v),F_u,F_v,E^i_{u,v})$, where each element of the sequence consists of a pair of nodes' indexes $\alpha(u)$ and $\alpha(v)$ for nodes $u$ and $v$, node attribute $F_u, F_v$, and the edge attribute $E^{(i)}_{u,v}$ for the edge at Step $i$. 
In edge-sequence-based generation, there are two ways to generate a unit $s_i$, with the first based on the assumption that $\alpha(u)$ and $\alpha(v)$ are mutually independent while the second assumes they are mutually dependent, with their details as follows. \\

\vspace{-0.1cm}
\noindent\textbf{Independency-based}. Goyal et al.~\cite{goyal2020graphgen} used depth first search (DFS) algorithm~\cite{yan2002gspan} as the ordering index function $\alpha(\cdot)$ to construct graph canonical index of nodes. The conditional distribution for generating each edge in graph $G$ can be formalized as \small
\begin{align}\nonumber
        &p(s_i|s_{<i})
        =p((\alpha(u),\alpha(v),F_u,F_v,E^i_{u,v})|s_{<i})\\
        &=p(\alpha(u)|s_{<i})p(\alpha(v)|s_{<i})p(F_u|s_{<i})p(F_v|s_{<i})p(E^i_{u,v}|s_{<i}),\label{eq:edge_sequence}
\end{align}\normalsize
where $s_{<i}$ refers to the already-generated edges and nodes. A customized long short-term memory (LSTM) is designed which consists of a transition state function $f_{\mathrm{trans}}$ for transferring the hidden state of the last step into that of the current step (in Eq.~\eqref{eq:trans}), an embedding function $f_{\mathrm{emb}}$ for embedding the already generated graph into latent representations (in Eq.~\eqref{eq:trans}), and five separate output functions for the above five distribution components (in Eq~\eqref{eq:trans} to Eq.~\eqref{eq:edge}). It is assumed that the five elements in one tuple are independent of each others, and thus the inference is as:\vspace{-0.3cm}

\small
\begin{align}
    &h_{G}^{(i)}=f_{\mathrm{trans}}(h_{G}^{(i-1)},f_{\mathrm{emb}}(s_{i-1}))\label{eq:trans} \\
    &\alpha(u) \sim Cat(\theta_{\alpha(u)});\quad \theta_{\alpha(u)}=f_{\alpha(u)}(h_{G}^{(i)});\quad\\&
    \alpha(v) \sim Cat(\theta_{\alpha(v)});\quad \theta_{\alpha(v)}=f_{\alpha(v)}(h_{G}^{(i)})\label{eq:node_i}\\
    & F_u \sim Cat(\theta_{F_u});\quad \theta_{F_u}=f_{F_u}(h_{G}^{(i)});\quad\\&
    F_v \sim Cat(\theta_{F_v});\quad \theta_{F_v}=f_{F_v}(h_{G}^{(i)})\\ \quad
    &E^i_{u,v} \sim Cat(\theta_{E^i_{u,v}});\quad \theta_{E^i_{u,v}}=f_{E^i_{u,v}}(h_{G}^{(i)}),\label{eq:edge}
\end{align}\normalsize
where $s_{i-1}$ refers to the generated tuple at Step $i-1$ and is represented as the concatenation of all the component representations in the tuple. $h_{G}^{(i)}$ is a graph-level LSTM hidden state vector that encodes the state of the graph generated so far at Step $i$. Given the graph state $h_{G}^{(i)}$, the output of five functions $f_{\alpha(u)}$, $f_{\alpha(v)}$, $f_{F_u}$, $f_{F_v}$, $f_{E_{u,v}}$ model the categorical distribution of the five components of the newly formed edge tuple, which are paramerized by five vectors $\theta_{\alpha(u)}$, $\theta_{\alpha(v)}$, $\theta_{F_u}$, $\theta_{F_v}$, $\theta_{E_{u,v}}$ respectively. 
Finally, the components of the newly formed edge tuple are sampled from the five learnt categorical distributions.\\

\vspace{-0.3cm}
\noindent\textbf{Dependency-based}. To further characterize the dependency between $\alpha(u)$ and $\alpha(v)$, Bacciu et al.~\cite{bacciu2019graph} assume the existence of node dependence in a tuple. This method deals with homogeneous graphs without considering the node/edge categories, by representing each tuple in the sequence as $s_i=(\alpha(u),\alpha(v))$ and formalizing the distribution as $p(s_i|s_{<i})=p(\alpha(u)|s_{<i})p(\alpha(v)|\alpha(u),s_{<i})$. Then, the first node is sampled in the same way as in Eq.~\eqref{eq:node_i}, while the second node in the tuple is sampled as follows:\vspace{-0.4cm}

\small
\begin{align}\vspace{-0.2cm}
 \alpha(v) \sim Cat(\theta_{\alpha(v)});\quad \theta_{\alpha(v)}=f_{\alpha(v)}(h_{G}^{(i)},g_{\mathrm{emb}}(\alpha(u))),
\end{align}\normalsize
where the function $g_{\mathrm{emb}}$ is used for embedding the index of the first generated node $u$ in the pair.

\subsubsection{Motif-sequence-based}

\textbf{General Framework}.
Motif-sequence-based methods~\cite{liao2019efficient,jin2018junction,podda2020deep,gu2019explore} represent a graph $G$ as a sequence of graph motifs, $Seq(G)=\{C_1,...,C_M\}$, where the block of nodes and edges that constitute each graph motif $C_i$ are generated at each step, as shown in Fig.~\ref{fig:sequence-generation}(c). A new graph-motif $C_i$ is generated in each step by conditioning on the current graph $G_i$ at Step $i$ and then it is connected to $G_i$.

A key problem in motif-based methods is how to connect the newly generated graph motif $C_i$ to 
$G_i$,
given that there are many potential ways to link two sub-graphs. These linking strategies are highly dependent on the definition of the graph motifs. For \textit{Domain-agnostic} graphs, given a predefined node ordering, the graph motifs are usually defined as a combination of consecutive nodes. This allows us to predict the associated edges of all the nodes in $C_i$ and connect it to $G_i$ based on these predictions. For \textit{Domain-specific} graphs, the motifs are usually defined and connected based on specific domain knowledge, such as chemical motifs for a task involving molecular structure generation.\\

\vspace{-0.1cm}
\noindent\textbf{Domain-agnostic-based}. This line of works is designed for generating general graphs without the need of domain expertise; it is similar to the collective-associated-edge-generation category under the line of  node-sequence generation by generating the adjacent vectors for each edge, such as GraphRNN~\cite{you2018graphrnn}, except for the generation of several nodes instead of one per step. As described in Section~\ref{sec:node-sequence}, a graph $G$ is represented as a sequence of node-based tuples as $G=\{s_1,...,s_N\}$, where $s_i=(v_i^\pi,A^\pi_{i,\cdot})$ is generated per step. Based on this node sequence, Liao et al.~\cite{liao2019efficient}(GRANs) regard every tuples consisting of $B$ recursive nodes as a graph motif $C_i$ and generates each block per step. In this way, the generated nodes in the new graph motif follow the ordering of the nodes in the whole graph and contain all the connection information of the existing and newly generated nodes. To formalize the dependency among the existing and newly generated nodes, GRANs proposes an MPNN-based model to generate the adjacent vectors. Specifically, for the $i$-th generation step, a graph $G_i$ is generated which contains the already-generated graph with $B\cdot(i-1)$ nodes and the edges among them, as well as the $B$ nodes in the newly generated graph motif. For these new $B$ nodes, edges are initially completely added to connect all of them with each other and the previous $B\cdot(i-1)$ nodes.
Then an MPNN-based graph neural network (GNN)~\cite{scarselli2008graph} on this augmented graph is used to update the nodes' hidden states by encoding the graph structure. After several rounds of message passing implementation, the node-level hidden states of both the existing and newly added nodes are used to infer the final distribution of the newly added edges as:
\begin{equation}\small 
    p(C_t|C_{<t})=\prod\nolimits_{B(t-1)<i\le B}\prod\nolimits_{1 \le j \le i}p(A^\pi_{i,j}|C_{<t})\\
\end{equation}\normalsize
where the Bernoulli distribution $p(A^\pi_{i,j}|C_{<t})$ is parameterized for modeling the edge existence through an MLP, which takes the node-level hidden states as input.\\

\vspace{-0.1cm}
\noindent\textbf{Domain-specific-based}. 
The definition of graph motifs and its connections can involve domain knowledge, such as in the situation of molecules generation (i.e., graph of atoms)~\cite{jin2018junction, podda2020deep}. Jin et al.~\cite{jin2018junction} propose the Junction-Tree-VAE by first generating a tree-structured scaffold over chemical substructures, and then combining them into a molecule with an MPNN. Specifically, a Tree Decomposition of Molecules algorithm~\cite{rarey1998feature} tailored for molecules to decompose the graph $G$ into several graph motifs $C_i$ is followed, and each $C_i$ is regarded as a node in the tree structure. The other way of defining the graph motifs is to leverage the breaking of retrosynthetically interesting chemical substructures (BRICS) algorithm~\cite{degen2008art}.
To generate a graph $G$, a tree is first generated and then converted into the final graph. The decoder for generating a $T$ consists of both \emph{topology prediction function} and \emph{label prediction function}. The topology prediction function models the probability of the current node to have a child, and the label prediction function models a distribution of the labels of all types of $C_i$. When reproducing a molecular graph $G$ that underlies the predicted junction tree $T$, since each motif contains several atoms, the neighboring motifs $C_i$ and $C_j$ can be attached to each other as sub-graphs in many potential ways. To solve this, a scoring function (e.g., measuring the validness of the potentially generated graph) over all the candidates graphs is proposed, and the optimal one that maximizes the scoring function is the final generated graph.


\subsubsection{Rule-sequence-based}
\label{sec:rule-based}

\textbf{General Framework}.
Several methods that have been proposed~\cite{dai2018syntax, kusner2017grammar} generate a sequence of production rules or commands to guide the graph construction process sequentially. This is usually the method of choice where the targeted graph has strong constraints or grammar that must be satisfied in order to construct a valid graph. For example, a molecule can not violate fundamental properties like charge conservation, which thus constrains the patterns available for the node types and edges of molecule graph. 
To ensure the validity of the generated graphs, graph generation is transformed by generating parse trees, which describe a discrete molecular structure utilizing context free grammar (CFG), while the parse tree itself can be further expressed as a sequence of rules based on a pre-defined order. 

Kusner et al.~\cite{kusner2017grammar} propose generating a parse tree that describes a discrete object (e.g. arithmetic expressions and molecule) by a grammar; they also proposed a graph generation method named GrammerVAE. An example of using the parse tree for molecule generation: to encode the parse tree, they decompose it into a sequence of production rules by performing a pre-ordered traversal on its branches from left-to-right, and then convert these rules into one-hot indicator vectors, where each dimension corresponds to a rule in the SMILES grammar. The deep convolutional neural network is then mapped into a continuous latent vector
$z$. While decoding, the continuous vector $z$ is passed through an RNN which produces a set of unnormalized log probability vectors (i.e., ``logits"). Each dimension of the logit vectors corresponds to a production rule in the grammar. The model generates the
parse trees directly in a top-down direction, by repeatedly expanding the tree with its production rules. The molecules are also generated by following the rules generated sequentially, as shown in Fig.~\ref{fig:sequence-generation}(d). Although the CFG provides a mechanism for generating syntactic-valid objects, it is still incapable of guaranteeing the model for generating semantic valid objects~\cite{kusner2017grammar}. To deal with this limitation, Dai et al.~\cite{dai2018syntax} propose the syntax-directed variational autoencoder (SD-VAE), in which a semantic restriction component is advanced to the stage of syntax-tree generator. This allows for a the generator with both syntactic and semantic validity.

\subsubsection{Comparison of different sub-categories}
\label{sec:2.1.5}
In this subsection, we compare the four categories of sequential-generating method from three aspects: (1) \textbf{Scalability}: time complexity determines the scalability of the graph generation methods. Node-sequence-based methods commonly have the time complexity of $O(N^2)$ when $N$ denotes to the number of nodes, while edge-sequence-based methods usually have the complexity of $O(|\mathcal{E}|)$. Thus, for sparse graphs where $N^2\gg |\mathcal{E}|$, edge-sequence-based methods are more scalable than node-sequence-based ones. The complexities of motif-sequence-based methods vary from $O(N^2)$ (e.g., for domain-agnostic type) to $O(N\cdot|C|)$ (e.g., for domain-specific type), where $|C|$ refers to the number of motifs. The complexity of rule-sequence-based methods usually linearly related to the number of rules in generating a graph;
(2) \textbf{Expressiveness}: the expressiveness of generation model relies on its power to model the complex dependency among the objects in the graph. Node-sequence and edge-sequence generation can capture the most sophisticated dependence, including node-node dependence, edge-edge dependence and node-edge dependence. While the motif-sequence-based methods are able to model the dependence between graph-motifs which capture the high-order relationships and global patterns. Rule-sequence-based methods can model the dependency between the operation rules to capture the semantic patterns in building a realistic graphs, which are usually difficult to directly learn  from the graph topology;
(3) \textbf{Application scenarios}: the selection of categories of sequential generating techniques for a specific application scenario depends on its sensitivity to validness and the accessibility of the generation rules. Node- and edge-sequence-based methods are suitable in generating realistic graphs without the domain expertise (e.g., the known rules, constrains or candidate motifs), such as the social and traffic networks.
Motif-sequence-based methods can partially guarantee the validness of the generated graph by selecting graph-motifs from the predefined valid motif candidates. Rule-sequence-based methods are more powerful in generating valid realistic graphs by following the correct grammar and constraints. Thus, the latter two types of methods are preferred in validness-sensitive applications, such as molecule generation and program modeling.

\subsection{One-shot generating}
One-shot generating methods learn to map each whole graph into a unified latent representation which follows some probabilistic distribution in latent space. Each whole graph can then be generated by directly sampling from this probabilistic distribution in one step. The core issue of these methods is usually how to jointly generate graph topology together with node/edge attributes.
Considering that the graph topology can usually be represented in terms of adjacency matrix and node attribute matrix, the typical solution is to learn the distribution of these two and generate them in one shot, which is categorized as \textit{Adjacent-matrix-based} generation. Learning the distribution of adjacency matrices is potentially expressive yet comes with inefficiency issue in both memory and time. To this end, \textit{Edge-list-based} methods learn the local patterns and hence is usually good at handling larger graphs with simpler global patterns.


\subsubsection{Adjacency-matrix-based}
\label{sec:adjacent-one-shot}

\textbf{General Framework}.
Adjacency-matrix-based methods build models to directly map the latent embedding $z$ to the output graph in terms of an adjacency matrix, generally with the addition of node/edge attribute matrices/tensors. Hence, how to best achieve an expressive and efficient mapping is the core challenge and there is usually a trade-off between them. Existing techniques are built upon popular deep neural network scenarios that are MLP-based, message-passing-based, invertible-transformation-based or transposed-convolution-based. MLP-based models are highly end-to-end, while message-passing-based approaches and transposed-convolution-based can explicitly model higher-order correlations in graphs. Invertible-transformation-based techniques more rigorously model invertible mappings but impose more limitations on the expressiveness.
\begin{figure}[htb]
    \centering
    \includegraphics[width=0.48\textwidth]{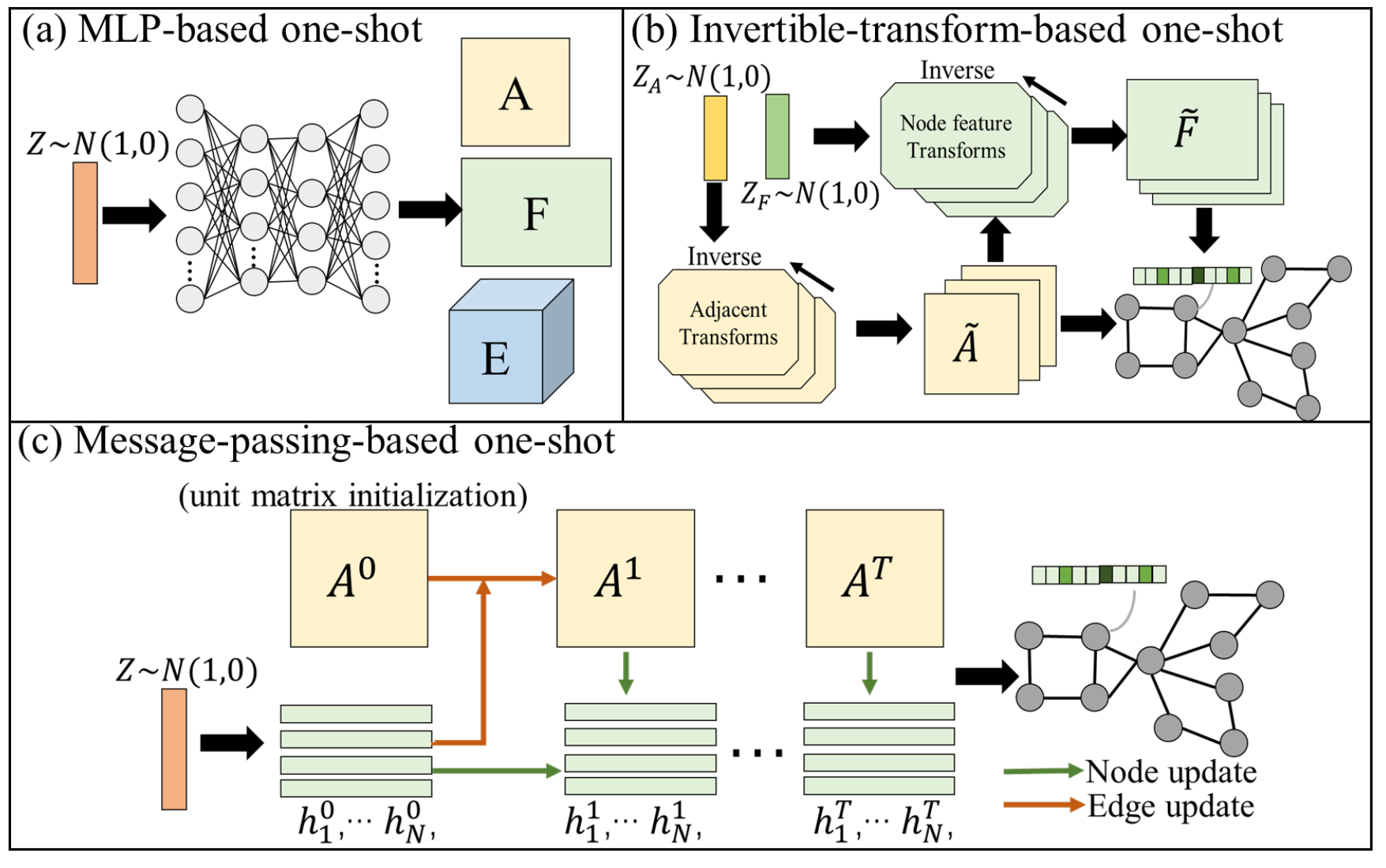}\vspace{-0.4cm}
    \caption{Schema for adjacent-matrix-based one-shot generation}
    \label{fig:one-shot}\vspace{-0.2cm}
\end{figure}

\noindent\textbf{MLP-based methods}.
Most of the one-shot graph generation techniques involves simply constructing the graph decoder $g(z)$ using MLP~\cite{simonovsky2018graphvae,ma2018constrained,anand2018generative,fan2019labeled,polsterl2019likelihood,de2018molgan}, where the models' parameters can be optimized under common frameworks such as VAE and GAN. The MLP-based models ingest a latent graph representation $z\sim p(z)$ and simultaneously output adjacent matrix $A^\pi$ and node attribute $F^\pi$, as shown in Fig.~\ref{fig:one-shot}(a).  
Specifically, the generator $g(z)$ takes D-dimensional vectors $z\in \mathbb{R}^D$ sampled from a statistical distribution such as standard normal distribution and outputs graphs. For each $z$, $g(z)$ outputs two continuous and dense objects: $\Tilde{A}^\pi$, which defines edge attributes and $\Tilde{F}^\pi$, which denotes node attributes
through two simple MLPs. Both $\Tilde{A}^\pi$ and $\Tilde{F}^\pi$ have a probabilistic interpretation since each node and edge attribute is represented with probabilities of categorical
distributions of types. To generate the final graph, it is required to obtain the discrete-valued objects $A^\pi$ and $F^\pi$ from $\Tilde{A}^\pi$ and $\Tilde{F}^\pi$, respectively. The existing works have two ways to realize this step detailed as follows.

In the first way, the existing works~\cite{simonovsky2018graphvae,anand2018generative,ma2018constrained} use sigmoid activation function to compute $A^\pi$ and $F^\pi$ during the training time. At test time, the discrete-valued estimate $A^\pi$ and $F^\pi$ can be obtained by taking edge- and node-wise argmax in $\Tilde{A}^\pi$ and $\Tilde{F}^\pi$. Alternatively, existing works~\cite{polsterl2019likelihood,de2018molgan,fan2019labeled} leverage categorical reparameterization
with the Gumbel-Softmax~\cite{jang2017categorical,maddison2016concrete}, which is to sample from a categorical
distribution during the forward pass (i.e., $F^\pi_i\sim Cat(\Tilde{F}^\pi_i)$ and $A^\pi_{ij} = Cat(\Tilde{A}^\pi_{ij}))$ and the original continuous-valued $\Tilde{A}^\pi$ and $\Tilde{F}^\pi$ in the backward pass. In this way, these methods can perform continuous-valued operations during the training procedure and do the categorical sampling procedure to finally generate $F$ and $A$.\\

\noindent\textbf{Message-passing-based methods}.
Message-passing-based methods generate graphs by iteratively refining the graph topology and node representations of the initialized graph through the MPNN. Specifically, based on the latent representation $z$ sampled from a simple distribution (e.g., Gaussian), we usually first generate an initialized adjacent matrix $A^0$ and the initialized node latent representations $H^0\in \mathbb{R}^{N\times L}$, where $L$ refers to the length of each node representation (here we omit the node ordering symbol $\pi$ for clarity). Then $A^0$ and $H^0$ are updated though MPNN into $A^1$ and $H^1$, which are the adjacent matrix and hidden states in the first intermediate layer, then another MPNN layer is applied to generate for the 2nd layer, etc. We can stack multiple such layers to explicitly characterize the higher-order correlation among nodes and edges. Each MPNN layer can be expressed as follows:\vspace{-0.2cm}

\vspace{-0.2cm}
\small
\begin{align}\nonumber
    &A^{l+1}_{i,j}=A^l_{i,j}+ReLu(\nu_1 A_{i,j}^l+\nu_2 h_{i}^l+\nu_3 h_{j}^l); \quad\\&
    h_i^{l+1}=h_i^l+ReLu(w_1 h^l_i+\sum\nolimits_{j}^{N}\eta_{i,j} w_2 h^l_j),\label{eq:mp2}
\end{align}\normalsize
where $v_1$, $v_2$, $v_3$, $w_1$ and $w_2$ are trainable parameters. We can stack multiple such layers to explicitly characterize the higher-order correlation among nodes and edges, which is also illustrated in Fig.~\ref{fig:one-shot}(c). Finally, after $T$ layers' updating, the outputs $A^T_{i,j}$ and $F^T_i$ are used to parameterize the categorical distributions of each edge and node, based on which each edge $A_{i,j}$ and node $F_i$ are generated through categorical sampling introduced above.  To learn the above generator, Existing methods leverage various learning frameworks such as VAE and GANs~\cite{bresson2019two,guarinodipol,flam2020graph}, or have a plain framework based on the score-based generation~\cite{niu2020permutation}.\\

\vspace{-0.2cm}
\noindent\textbf{Invertible-transform-based methods}.
Flow-based generative methods can also do one-shot generation, by a unique invertible function between graph $G$ and the latent prior $z$ sampling from a simple distribution (e.g., Gaussian), as shown in Fig.~\ref{fig:one-shot}(b). Concretely, based on vanilla flow-based learning techniques introduced in Section~\ref{sec:flow-based}, special forward transformation $G\xrightarrow{}z$ and backward transformation $z\xrightarrow{}G$ needs to be designed.

Madhawa et al.~\cite{madhawa2019graphnvp} propose the first flow-based one-shot graph generation model called GraphNVP. 
To get $z=(z^F, z^A)$ from $G=(A,F)$ in the \textit{forward transformation}, they first convert the discrete variable $A$ and $F$ into continuous variable $A'$ and $F'$ by adding real-valued noise), which is known as \emph{dequantization}. Then two types of reversible affine coupling layers: adjacency coupling layers and node attribute
coupling layers are utilized to transform the adjacency matrix $A'$ and the node attribute matrix $F'$ into latent representations $z_A$ and $z_F$, respectively. The $l$th reversible coupling layers are designed as follows:\vspace{-0.2cm}

\vspace{-0.2cm}
\small
\begin{align}
    &z^{l}_F[i]=z^{l-1}_F[i]\odot \mathrm{exp}(s_F(z^{l-1}_F[i],A))+t_F(z^{l-1}_F[i],A)\label{eq:graphnvp1}\\
    &z^{l}_A[i,j]=z^{l-1}_A[i,j]\odot \mathrm{exp}(s_A(z^{l-1}_A[i,j]))+t_A(z^{l-1}_A[i,j])\label{eq:graphnvp2}
\end{align}\normalsize
where $z^{0}_F=X'$ and $z^{0}_A=A'$. $z^{l}_F[i]$ refers to the $i$th entry of $z^{l}_F$; $\odot$ denotes element-wise multiplication. Functions $s_A(\cdot)$ and $t_A(\cdot)$ stand for scale and translation operations which can be implemented based on MPNN, and $s_F(\cdot)$, $t_F(\cdot)$ can be implemented based on MLP networks. 
To get $G=(F,A)$ from $z=(z_F, z_A)$ in the \textit{backward transformation}, the reversed operation is conducted based on the above forward transformation operation in Eq.~\eqref{eq:graphnvp1} and~\eqref{eq:graphnvp2}. 
Next a probabilistic feature matrix $\Tilde{F}$ is generated given the sampled $z_F$ and the generated adjacency matrix $A$ through a sequence of inverted node attribute coupling layers. Likewise, the node-wise argmax of $\Tilde{F}$ is used to get discrete feature matrix $F$. \\


\vspace{-0.1cm}
\noindent\textbf{Transposed-convolution-based methods}. One typical type of graph decoder in the one-shot-generation techniques is constructed based on the transposed convolution neural networks~\cite{guo2022deep}. The process is about generating the adjacent matrix of graph by taking the node latent representation vectors as input. The transposed-convolution-based decoder consists of a node transposed convolution layer and several edge transposed convolution layers. 

The node transposed convolution layer is used to decode the edge representations of the graph based on the node embedding. For example, after a node transposed convolution layer, the edge representations $E_{i,j}$ between node $v_i$ and node $v_j$ can be computed as:\vspace{-0.2cm}

\small
\begin{equation}
\label{eq:n2edeconv_direct}
 E_{i,j}=\sum\nolimits^{L}_{m=1}(\sigma (H_i^{m}\Bar{\mu}_j)+\sigma(H_j^{m}\Bar{\nu}_i)),
\end{equation}\normalsize
where $\sigma (H_i^{m}\Bar{\mu}_j)$ means the transposed convolution contribution of node $v_i$ to its potential edge $E_{i,j}$, which is made by the $m$-th entry of its node representations, and $\Bar{\mu}_j$ represents one entry of the transposed convolution filter vector $\Bar{\mu}\in\mathbb{R}^{N \times 1}$ that is related to node $v_j$. $L$ refers to the length of the node representation.

Several edge transposed convolution layers are recursively applied to decode the latent edge representations from the upper layer back to those of the lower layer.
Thus, $E^{l}_{i,j}$ between node $v_i$ and node $v_j$ in the $(l+1)$th layer is computed as:\vspace{-0.3cm}

\small
\begin{equation}
\label{eq:e2edeconv_direct}
 E^{l+1}_{i,j}=\sigma (\Bar{\phi}^{l}_j\sum\nolimits_{k_1=1}^{N}E^{l}_{i,k_1}S_{k_1})+\sigma(\Bar{\psi}^{l}_i\sum\nolimits_{k_2=1}^{N}E^{l}_{k_2,j}S_{k_1}),
\end{equation}\normalsize 
where $\Bar{\phi}^{l}_j\sum_{k_1=1}^{N}E^{l}_{i,k_1}S_{k_1}$ can be interpreted as the decoded contribution of node $v_i$ to its related edge representations $E^{l+1}_{i,j}$, and $\Bar{\phi}^{l}_j$ refers to the element of transposed convolution filter vector that is related to node $v_j$. $\sigma$ refers to the activation functions. 

\subsubsection{Edge-list-based}
\textbf{General Framework}.
This category typically requires a generative model that learns edge probabilities, where all the edges are generated independently. These methods are usually applied when learning from one large-scale graph to generate a new one using the existing nodes. The general pipeline is composed of two main steps. A score is calculated for each edge (i.e., pair of nodes) to estimate the edge probability, after which the edges can be sampled.

In terms of how the edge probabilities are generated, existing works are further categorized as either \emph{random-walk-based}~\cite{bojchevski2018netgan,zhang2019stggan,gamage2020multi,carida2019can} or \emph{node-similarity-based}~\cite{kipf2016variational,grover2019graphite,zou2019encoding,liu2019graph,salha2019gravity}. \emph{Node-similarity-based} models calculate the edge probability based on the similarity of each pair of node representations learnt from graphs, while \emph{random-walk-based} methods estimate each edge probability by calculating the edge frequency for a large set of random walks generated by sampling from their distributions learnt from graphs. \\

\vspace{-0.2cm}
\noindent\textbf{Random-walk-based}.
This type of methods generate the edge probability based on a score matrix, which is calculated by the frequency of each edge that appears in a set of generated random walks.  NetGAN~\cite{bojchevski2018netgan} is proposed to mimic the large-scale real-world networks. Specifically, at the first step, a GAN-based generative model is used to learn the distribution of random walks over the observed graph, and then it generates a set of random walks. At the second step, a score matrix $S\in\mathcal{R}^{N\times N}$ is constructed, where each entry denotes the counts of an edge that appears in the set of generated random walks. At last, based on the score matrix, the edge probability matrix $\Tilde{A}$ is calculated as $\Tilde{A}_{i,j}=S_{i,j}/\sum_{u,v}^NS_{u,v}$, which will be used to generate individual edge $A_{i,j}$, based on efficient sampling processes.   

Following this, some works propose improving the NetGAN, by changing the way to choose the first node in starting a random walk~\cite{carida2019can} or learning spatial-temporal random walks for spatial-temporal graph generation~\cite{zhang2019stggan}. Gamage et al.~\cite{gamage2020multi} generalize the NetGAN by adding two motif-biased random-walk GANs. The edge probability is thus calculated based on the score matrices from three sets of random walks (i.e. $S^{(1)}$, $S^{(2)}$, and $S^{(3)}$) that are generated from the three GANs. To sample each edge, one view $S^{(k)}$ is randomly selected from the three scores matrices. Based on $S^{(k)}$, edge probability $\Tilde{A}_{i,j}$ is calculated as $\Tilde{A}_{i,j}=S^{(k)}_{i,j}/\sum_{u,v}^NS_{u,v}$.  \\

\noindent\textbf{Node-similarity-based}.
These methods generate the edge probability based on pairwise relationships between the given or sampled nodes' embedding (as in \cite{kipf2016variational}).
Specifically, the probability adjacent matrix $\Tilde{A}$ is generated given the node representations $Z \in \mathcal{R}^{N\times L}$, where $Z_i\in \mathcal{R}^{L}$ refers to the latent representation for node $v_i$. $\Tilde{A}$ will be used to generate individual edge $A_{i,j}$ , based on efficient sampling processes. Existing methods differ on how to calculate $\Tilde{A}$.

Several works~\cite{kipf2016variational,grover2019graphite,zou2019encoding} compute $\Tilde{A}_{i,j}$ based on the inner-product operations of two node embedding $Z_i$ and $Z_j$. This reflects the idea that nodes that are close in the embedding space should have a high probability of being connected. These works require a setting where node set is pre-defined and the node attribute $F$ is known in advance. Specifically, by first sampling node latent representation
$Z_i$ from the standard normal distribution, Kipf et al.~\cite{kipf2016variational,grover2019graphite} calculate the probability adjacent matrix as $\Tilde{A}=\mathrm{Sigmoid}(ZZ^T)$.
The adjacent matrix $A$ is then sampled from $\Tilde{A}$ which parameterizes the Bernoulli distribution of the edge existence, as similar to work in \cite{zou2019encoding}. 

Other works~\cite{liu2019graph,salha2019gravity} compute $\Tilde{A}_{i,j}$ by measuring the closeness of two node,representations with $\ell_2$ norm. Liu et al.~\cite{liu2019graph} propose a decoder for calculating $\Tilde{A}_{i,j}$ as:\vspace{-0.2cm}

\small
\begin{equation}\vspace{-0.15cm}
    \Tilde{A}_{i,j}=1/(1+exp(C(\parallel Z_i-Z_j\parallel^2_2-1))),
\end{equation}\normalsize
where $C$ is called a temperature hyperparameter. Salha et al.~\cite{salha2019gravity} propose a gravity-inspired decoding schema as:\vspace{-0.2cm}

\small
\begin{equation}\vspace{-0.1cm}
  \Tilde{A}_{i,j}=\mathrm{Sigmoid}(m_j-log\parallel Z_i-Z_j\parallel^2_2),  
\end{equation}\normalsize
where $m_j$ is the gravity scale of node $v_j$ learned from the input graph by its featured encoder.

\subsubsection{Comparison of different sub-categories}
\label{sec:2.3.2}
In this subsection, we compare two aspects of the two different types of one-shot method: (1) \textbf{Time complexity}: Both adjacent-matrix-based and node-similarity-based edge-list generation have a complexity of $O(N^2)$ since they need to consider every pairs of $N$ nodes in the graph. Random-walk-based edge-list generation is more scalable, as here the edges are sampled based on the edge probability, which is determined by the edge frequency in a set of generated random walks; and (2) \textbf{Application scenarios}:  Since  adjacent-matrix-based methods can handle global patterns with high expressiveness and minimum time consumption, these types of methods are widely used for small graphs (i.e., graphs with less than 1,000 nodes) whose global patterns are important, such as molecules and proteins. Edge-list-based methods, on the other hand, are efficient when it comes to generating large graphs whose local patterns are important, such as social networks and citation networks. 


\section{Conditional Deep Generative Models for Graph Generation}
\label{sec:four}
The goal of conditional deep graph generation is to learn a conditional distribution $p_{\mathrm{model}}(G|y)$ based on a set of observed realistic graphs $G$ along with their corresponding auxiliary information, namely a condition $y$. The auxiliary information could be category labels, semantic context, graphs from other distribution spaces, etc.

Compared with unconditional deep graph generation, in addition to the challenge in generating graphs, conditional generation needs to consider how to extract the features from the given condition and integrate them into the generation of graphs. Thus, to systematically introduce the existing conditional deep graph generative models, we mainly focus on describing how these methods deal with the conditions. Since the conditions could be any form of auxiliary information, they are categorized into three types, including \textbf{graphs}, \textbf{sequence}, and \textbf{semantic context}, shown as the yellow parts of the taxonomy tree in Fig.~\ref{fig:tree}.\vspace{-0.2cm}

\begin{table*}[htb]
    \centering
    \caption{Deep Generative-based Methods for Conditional Graph Generation}\vspace{-0.3cm}  
    \begin{tabular}{|l|l|l|l|}
    \toprule\hline
         \multicolumn{2}{|l|}{Conditioning objects}& Techniques of encoding conditions&References \\\hline
         \multirow{3}{*}{Graphs}&Edge Transformation&Adjacent-based edge convolution&\cite{guo2022deep,do2019graph,gao2018local,zhou2019misc}\\\cline{2-4}
         ~&\multirow{2}{*}{Node-edge Co-transformation}&Embedding-based&\cite{kaluzaneural,sun2019graph,jin2018learning,maziarka2020mol}\\\cline{3-4}
         ~&~&Editing-based&\cite{you2018graph,zhou2019optimization,jin2020composing}\\\hline      
         \multicolumn{2}{|l|}{Sequence} & RNN-based encoding&\cite{chen2018sequence,wang2018neural,yang2020learn,liu2015dynamic}\\\hline
         \multicolumn{2}{|l|}{Context Semantics} &Concatenation with latent representation&\cite{yang2019conditional,li2018learning,jonas2019deep,li2018multi}\\\hline
         \hline
    \end{tabular}\vspace{-0.2cm}
    \label{tab:my_label}
\end{table*}

\vspace{-0.2cm}
\subsection{Conditioning on graphs}
\label{sec:graph translation}
The problem of deep graph generation conditioning on another graph is also called as deep graph transformation (also known as deep graph translation)~\cite{guo2022deep}. It aims at translating an input graph $G_S$ in the source domain to the corresponding output graphs $G_T$ in the target domain. Considering the entities that are being transformed during the translation process, there are two categories of works in the domain of deep graph generation conditioning on graphs: \textit{edge transformation} and \textit{node-edge-co-transformation}\footnote{Purely node and edge attribute transformation have been handled in node classification or link prediction task by typical GNNs~\cite{kipf2016semi,gilmer2017neural}, thus are not the focus of our survey.}. 

\subsubsection{Edge transformation}
\textbf{Overall Problem Formulation}.
The problem of edge transformation is to generate the graph topology and edge attributes of the target graph conditioning on the input graph.
It requires the edge set $\mathcal{E}$ and edge attributes $E$ to change while the graph node set and node attributes are fixed during the translation process as: $\mathcal{T}: G_S(\mathcal{V},\mathcal{E}_S,F,E_S)\xrightarrow{} G_T(\mathcal{V},\mathcal{E}_T,F,E_T)$. The edge transformation problem has a wide range of real-world applications, such as modeling chemical reactions~\cite{you2018graph}, protein folding~\cite{anand2018generative} and malware cyber-network synthesis~\cite{guo2022deep}. Existing works adopt different frameworks to model the translation process.

Some works utilize the encoder-decoder framework by learning abstract latent representation of the input graph through the encoder and then generating the target graph based on these hidden information through the decoder~\cite{guo2022deep,gao2018local}. For exampl, Guo et al.~\cite{guo2022deep} propose a GAN-based model for graph topology transformation. The proposed GT-GAN consists of a graph translator and a conditional graph discriminator. The graph translator includes two parts: graph encoder and graph decoder. A graph convolution neural net~\cite{kawahara2017brainnetcnn} is extended to serve as the graph encoder in order to embed the input graph into node-level representations while a new graph deconvolution net is used as the decoder to generate the target graph. 

Zhou et al.~\cite{zhou2019misc} propose modeling the underlying distribution of graph structures of the input graph at different levels of granularity, and then “transferring” such hierarchical distribution from the graphs in the source domain to a unique graph in the target domain. The input graph is characterized as several coarse-grained graphs by aggregating the strongly coupled nodes with a small algebraic distance to form coarser nodes. Overall, the framework can be separated into three stages. At the first step, the coarse-grained graphs at $K$ levels of granularity are constructed from the input graph adjacent matrix $A_S$. The adjacent matrix of the coarse-grained graph $A_S^{(l)}\in \mathbb{R}^{N^{(l)}\times N^{(l)}}$ at the $k$th layer is defined as:\vspace{-0.2cm}

\small
\begin{equation}\vspace{-0.15cm}
 A_S^{(k)}={P^{(k-1)}}^T... {P^{(1)}}^T A_S P^{(1)}... P^{(k-1)},
\end{equation}\normalsize
where $P^{(k)}\in \mathbb{R}^{N^{(l)}\times N^{(l)}}$ is a coarse-grained operator for the $k$th level and $N^{(l)}$ refers to the number of nodes of the coarse-grained graph at level $l$. In the next stage, each coarse-grained graph at each level $k$ will be reconstructed back into a fine graph adjacent matrix $A_T^{(k)}\in \mathbb{R}^{N^{(l)}\times N^{(l)}}$ as:\vspace{-0.2cm}

\small
\begin{equation}\vspace{-0.15cm}
   A_T^{(k)}={R^{(1)}}^T...{R^{(k-1)}}^T A_S^{(k)} R^{(k-1)}...R^{(1)},
\end{equation}
where $R^{(k)}\in \mathbb{R}^{N^{(l)}\times N^{(l)}}$ is the reconstruction operator for the $k$th level. Thus all the reconstructed fine graphs at each layer are in the same scale. Finally, these graphs are aggregated into a unique one by a linear function to get the final adjacent matrix as follows: $A_T=\sum^K_{k=1}w^{k} A_T^{(k)}+b^{k}$, where $w^{k}\in\mathbb{R}$ and $b^{k}\in\mathbb{R}$ are weights and bias.


\subsubsection{Node-edge co-transformation}
\textbf{Overall Problem Formulation}.
The problem of node-edge co-transformation (NECT) is generating the node and edge attributes of the target graph conditioning on those of the input graph. It requires that both the nodes and edges can vary during the transformation process between the source graph and the generated target graph as follows: $\mathcal{T}: G_S(\mathcal{V}_S,\mathcal{E}_S,F_S,E_S)\xrightarrow{} G_T(\mathcal{V}_T,\mathcal{E}_T,F_T,E_T)$. In terms of the techniques on how the input graph is assimilated to generate the target graph, there are two categories: one is embedding-based and the other is editing-based.\\
\begin{figure}[htb]\vspace{-0.3cm}
    \centering
    \includegraphics[width=0.5\textwidth]{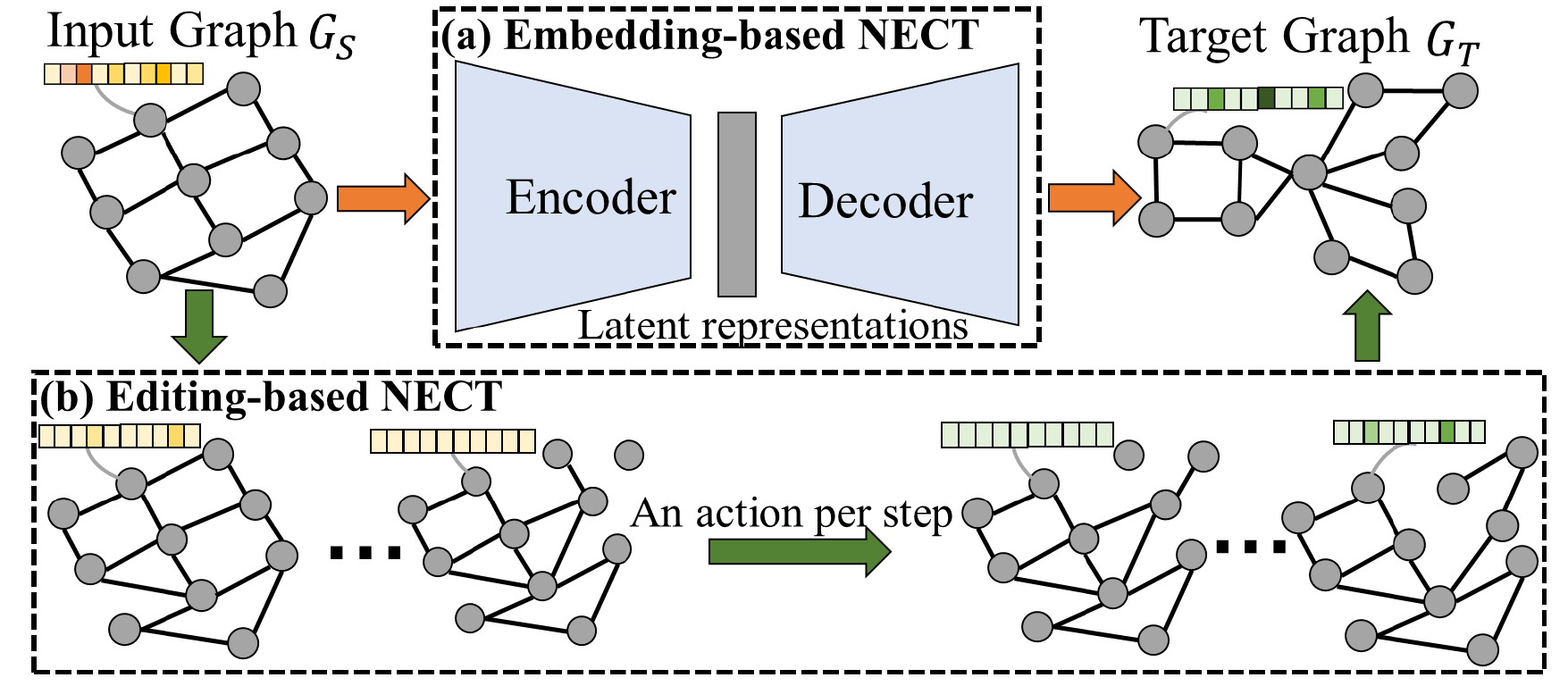}\vspace{-0.2cm}
    \caption{Embedding-based NECT vs Editing-based NECT}
    \label{fig:NECT}\vspace{-0.3cm}
\end{figure}

\noindent\textbf{Embedding-based NECT}.
The embedding-based NECT normally encodes the source graph into latent representations containing higher-level rich information of the input graph by an encoder, which is then decoded into the target graph by a decoder, as shown in Fig.~\ref{fig:NECT}(a)~\cite{kaluzaneural,sun2019graph,jin2018learning,maziarka2020mol,jin2020composing}. These methods are usually based on conditional VAEs~\cite{sohn2015learning} and conditional GANs~\cite{mirza2014conditional}.

Kaluze et al.~\cite{kaluzaneural} propose exploring the latent spaces of directed acyclic graphs (DAGs) and develops a neural network-based DAG-to-DAG translation model, where both the domain and the range of the target function are DAG spaces. 
The encoder $M_{\mathrm{encode}}$ is borrowed from the deep-gated DAG recursive neural network (DG-DAGRNN)~\cite{amizadeh2018learning}, which generalizes stacked RNNs on sequences to DAG structures. Each layer of the DG-DAGRNN consists of gated recurrent units (GRUs), which are repeated for each node $v_i$. The encoder outputs an embedding $h= M_{\mathrm{encode}}(G_S)$, which serves as the input
of the DAG decoder. The decoder follows the local-based node-sequential generation style as described in Section~\ref{sec:node-sequence}. Specifically, first, the number of nodes $N$ of the target graph is predicted by an MLP with the input of $h$. Also, the hidden state of the target graph is initialized with $h$. Then at each step, a node $v_i$ as well as its corresponding edge set $\{e_{i,j}\}_{j<i}$ are generated based on the hidden state at each step until an end node is added to the graph or the number of nodes exceeds a predefined threshold. Following this, a general graph-to-graph model~\cite{sun2019graph} is proposed by first formalizing the graph into a DAG without loss of information and utilize recurrent based model to translate this DAG. They embeds the topology of the input graph into the node representations by exerting a topology constraint, which results in a topology-flow encoder. Their decoder follows the same node sequential-based generation as proposed by You et al.~\cite{you2018graphrnn}. 
There are also some embedding-based graph translation methods that represent the graph as a set of graph motifs, which are usually targeted for the task of molecule optimization~\cite{jin2018learning,maziarka2020mol}. \\


\noindent\textbf{Editing-based NECT}.
Different from the encoder-decoder framework, Editing-based NECT directly modifies the input graph iteratively to generate the target graphs~\cite{you2018graph,zhou2019optimization,guo2019deep}, as shown in Fig.~\ref{fig:NECT}(b). There are two ways to realize the process of editing the source graph. One is utilizing an RL agent to sequentially modify the source graph based on a formulated Markov decision process\cite{you2018graph,zhou2019optimization} as described in Section~\ref{sec:RL}. The modification at each step will be selected from the defined action set, including ``add node'', ``add edge'', ``remove bonds'' et al. The other is to update nodes and edges from the source graph synchronously in a one-shot manner through the MPNN using several iterations~\cite{guo2019deep}. 

You et al.~\cite{you2018graph} propose the graph convolutional policy network (GCPN), a general graph convolutional network based model for goal-directed graph generation through reinforcement learning. The model is trained to optimize the domain-specific property of the source molecule through policy gradient, and acts in an environment that incorporates domain-specific rules. They define a distinct, fixed-dimension and homogeneous action space amenable to reinforcement learning, where an action is analogous to link prediction. Specifically, they first define a set of scaffold sub-graphs $\{C_1,...,C_s\}$ based on the source graph. This set acts as a sub-graph vocabulary that contains the sub-graphs to be added into the target graph during graph generation. Given the modified graph $G_t$ at step $t$, they define the corresponding extended graph as $G_t\cup C_i$. Based on this definition, an action can either correspond to connecting a new sub-graph $C_i$ to a node in $G_t$ or connecting existing nodes within graph $G_t$.

Guo et al.~\cite{guo2019deep} propose another way which edits the source graph iteratively, through the generation process extended from \emph{MPNN-based adjacency-based one-shot method} in Section~\ref{sec:adjacent-one-shot} and Fig.~\ref{fig:NECT}(c), which conducts the generation on both the node and edge attributes. The transformation process is modeled by several stages and each stage generates an immediate graph. Specifically, at each stage $t$, there are two paths, namely node translation and edge translation paths. In node translation path, an MLP-based \emph{influence-function} is used for calculating the influence $I^{(t)}_{i}$ on each node $v_{i}$ from its neighboring nodes, and another MLP-based \emph{updating-function} is used for updating the node attribute as $F_i^{(t)}$ with the input of influence $I^{(t)}_{i}$. The edge translation path is constructed in the same way as the node translation path, where each edge is generated by the influence from its adjacent edges. In addition, to capture and maintain the consistent between nodes and edges in the generated graph, a spectral-based regularization is enforced into the final optimization objective. 

\subsubsection{Comparison of different sub-categories}
\label{sec:3.1.3}
In this sub-section, we compare the two categories of methods in dealing with the node-edge-co-transformation (NECT). Since the comparison between different generating techniques is provided in Section~\ref{sec:three}, here we focus on the discussion regarding the relationship between the input and target graphs from three aspects: (1)~\textbf{Patterns captured from input graphs}: embedding-based NECT can capture the influences from the global patterns (e.g., density or molecule energy) of the input graphs onto the target graph with a graph-level latent representation. While the editing-based NECT has the advantage in modeling the influences from the local patterns (e.g., ``hub" node or ring structure) of the input graphs onto the target graphs;  (2)~\textbf{Interpretability}: editing-based NECT provides a more interpretable way by explicitly showing the transformation in a step-by-step fashion from the input to target graphs, which is more suitable to applications which rely on high-level confidence; While embedding-based NECT roughly connect the input and target graphs with a latent embedding which can not be semantically explained. (3)~\textbf{Application scenarios}: embedding-based NECT is capable of modeling the transformation with major and sophisticated changes from the input to target graphs, while editing-based NECT is more suitable to deal with the transformation with only small change, considering the efficiency.   

\subsection{Conditioning on sequence}
The problem of deep graph generation conditioning on a sequence can be formalized as the deep sequence-to-graph transformation problem. It aims to generate the target graph $G_T$ conditioning on an input sequence $X$. The deep sequence-to-graph problem is usually observed in domains such as NLP~\cite{chen2018sequence,wang2018neural} and time series mining~\cite{yang2020learn,liu2015dynamic}. 

The existing methods handle semantic parsing task~\cite{chen2018sequence,wang2018neural} by transforming a sequence-to-graph problem into a sequence-to-sequence problem and utilizing the classical RNN-based encoder-decoder model to learn this mapping. Chen et al.~\cite{chen2018sequence, wang2018neural} propose a neural semantic parsing approach named \textit{Sequence-to-Action}, which models semantic parsing as an end-to-end semantic graph generation process. Given a sentence $X = \{x_1,...,x_m\}$, the \textit{Sequence-to-Action} model generates a sequence of actions $Y = \{y_1,..,y_m\}$ for constructing the correct semantic graph. A semantic graph consists of nodes (including variables, entities, types) and edges (semantic relations), with some universal operations (e.g., argmax, argmin, count, sum, and not). To generate a semantic graph, they define six
types of actions: \textit{Add Variable Node}, \textit{Add Entity Node}, \textit{Add Type Node}, \textit{Add Edge}, \textit{Operation Function} and \textit{Argument Action}. In this way, the generated parse tree is represented as a sequence, and the sequence-to-graph problem is transformed into a sequence-to-sequence problem. Then the attention-based sequence-to-sequence RNN model~\cite{bahdanau2015neural} with an encoder and decoder is utilized, where the encoder converts the input sequence $X$ to a sequence of context sensitive vectors $\{b_1,...,b_m\}$ using a bidirectional RNN and a classical attention-based decoder generates action sequence $Y$.

Other methods handle the problem of Time Series Conditioned Graph Generation~\cite{yang2020learn,liu2015dynamic}: given an input multivariate time series, the aim is to infer a target relation graph to model the underlying interrelationship between the time series and each node. Yang et al.~\cite{yang2020learn} explore GANs in the conditional setting and propose the novel model of time series conditioned graph generation-generative adversarial networks (TSGG-GAN) for time series conditioned graph generation. Specifically, the generator in a TSGG-GAN adopts a variant of recurrent neural network called simple recurrent units (SRU)~\cite{lei2018simple} to extract essential information from the time series, and uses an MLP to generate the directed weighted graph.

\subsection{Conditioning on semantic context}
The problem of deep graph generation conditioning on semantic context aims to generate the target graph $G_T$ conditioning on an input semantic context, which can be usually represented as additional meta-features. The semantic context can refer to the category, label, modality or any additional information that can be intuitively represented as a vector $C$. The main issue is deciding where to concatenate or embed the condition representation into the generation process. As a summary, the conditioning information can be added in terms of one or multiple of the following
modules: (1) the node state initialization module, (2) the message passing process for MPNN-based decoding, and (3) the conditional distribution parameterization for sequential generating. 

Yang et al.~\cite{yang2019conditional} propose a novel unified model of graph variational generative adversarial nets, where the conditioning semantic context is inputted into the node state initialization module. Specifically, in the generation process, they first model the embedding $Z_i$ of each node with separate latent distributions. Then, a conditional graph VAE (CGVAE) can be directly constructed by concatenating the condition vector $C$ to 
each node latent representation $Z_i$ to get the updated node latent representation $\hat{Z}_i$. Thus, the distribution of the individual edge $A_{i,j}$ is assumed as a Bernoulli distribution, which is parameterized by the value $\hat{A}_{i,j}$ and is calculated as $\hat{A}_{i,j}=\mathrm{Sigmoid}(f(\hat{Z}_i)^T f(\hat{Z}_j))$,
where $f(\cdot)$ is constructed by a few fully connected layers. Li et al.~\cite{li2018learning} propose a conditional deep graph generative model that adds the semantic context information into the initialized latent representations $Z_i$ at the beginning of the decoding process.

Li et al.~\cite{li2018multi} add the context information $C$ into the message passing module in its MPNN-based decoding process. Specifically, they parameterize the decoding process as a Markov process and generate the graph by iteratively refining and updating from the initialized graph. At each step $t$, an action is conducted based on the current node hidden states $H^t=\{h_1^t,...,h_N^t\}$. To calculate $h_i^t\in\mathbb{R}^L$ ($L$ denotes the length of the representation) for node $v_i$ in the intermediate graph $G_t$ after each updating of the graph, they utilize message passing network with node message propagation. Thus the context information $C\in\mathbb{R}^K$ is added to the operation of the MPNN layer as follows:\vspace{-0.3cm}

\small
\begin{equation}\vspace{-0.2cm}
    h_i^t=Wh_i^{t-1}+\Phi\sum\nolimits_{v_j\in N(v_j)}h_j^{t-1}+\Theta C,
\end{equation}\normalsize
where $W\in{R}^{L\times L}$, $\Theta\in{R}^{L\times L}$ and $\Phi\in{R}^{K\times L}$ are all learnable weights vectors and $K$ denotes the length of the semantic context vector.
\vspace{-0.2cm}

\section{Evaluation Metrics for Deep Graph Generation}
\label{sec:five}
Evaluating the generated graphs as well as the learnt distribution of graphs are challenging and critical tasks for deep generative models in graph generation problem due to two major reasons: 1) Different from conventional prediction problems where merely deterministic predictions need to be evaluated, deep graph generation requires the evaluation of the learnt distributions. 2) Graph structured data is much more difficult to evaluate than simple data with matrix/vector structures or semantic data such as images and texts. Thus, we summarize the typical evaluation metrics in evaluating deep generative models for graph generation as shown in Figure~\ref{tab:metric}. We first provide the metrics that can be used both for unconditional and conditional deep graph generation, and then introduce the metrics that are specially designed for conditional deep graph generation. \vspace{-0.2cm}
\small
\begin{table}[htb]
    \centering
    \caption{Evaluation metrics for deep generative-based methods}\vspace{-0.3cm}  
    \begin{tabular}{|l|l|l|}
    \hline\hline
         \multicolumn{2}{|l|}{\textit{Type}}& \textit{Evaluation feature}  \\\hline
         \multirow{7}{*}{General}&\multirow{2}{*}{Statistics-based}&Average KLD\\\cline{3-3}
         ~&~& MMD\\\cline{2-3}
         ~&\multirow{2}{*}{Classifier-based}&Accuracy-based\\\cline{3-3}
         ~&~& FID-based\\\hline
        ~&\multirow{3}{*}{Intrinsic-quality-based}&Validity\\\cline{3-3}
         ~&~& Uniqueness\\\cline{3-3}
        ~&~& Novelty\\\hline
        \multirow{2}{*}{Condition-specialized} & \multicolumn{2}{|l|}{Graph property-based}\\\cline{2-3}
        ~& \multicolumn{2}{|l|}{Mapping-relationship-based}\\\hline
         \hline
    \end{tabular}\vspace{-0.2cm}
    \label{tab:metric}
\end{table}\normalsize

\subsection{General evaluation for deep graph generation}
To evaluate the quality of the generated graphs, existing literature covers three categories of evaluation metrics, namely statistics-based, classifier-based, and intrinsic-quality-based evaluations. The first two evaluation categories require comparison between the generated graph set and real graph set, while the intrinsic-quality evaluation directly measure the properties of the generated graph.

\subsubsection{Statistics-based}
In statistics-based evaluation, the quality of the generated graphs is accessed by computing the distance between the graph statistic distribution of real graphs and generated graphs. We first introduce seven typical graph statistics that measure different properties of graphs and, thereafter introduce the metrics that measure the distance between two distributions regarding different graph statistics.

There are seven typical graph statistics that are used in existing literature, which are summarized as follows:
(1) \emph{Node degree distribution}: the empirical node degree
distribution of a graph, which could encode its local connectivity patterns.
(2) \emph{Clustering coefficient distribution}: the empirical clustering coefficient distribution of a graph. Intuitively, the clustering coefficient of a node is calculated as the ratio of the potential number of triangles the node could be part of to the actual number of triangles the node is part of.
(3) \emph{Orbit count distribution}; the distribution of the counts of node 4-orbits of a graph. Intuitively, an orbit count specifies how many of these 4-orbits substructures the node is part of. This measure is useful in understanding if the model is capable of matching higher-order graph statistics, as opposed to node degree and clustering coefficient, which represent measures of local (or close to local) proximity. 
(4) \emph{Largest connected component}: the size of the largest connected component of the graphs.
(5) \emph{Triangle count}: the number of triangles counted in the graph.
(6) \emph{Characteristic path length}: the average number of steps along the shortest paths for all node pairs in the graph.
(7) \emph{Assortativity}: the Pearson correlation of degrees of connected nodes in the graph.

The first three graph statistics are about distributions of each graph and are always represented as a vector, while the last four graph statistics are represented as scalar values of each graph. Therefore, to evaluate the distance between two sets of graphs in terms of the above distribution statistics, two major metrics are usually utilized in existing literature, which are introduced as follows.

\emph{Average Kullback-Leibler Divergence}. 
Considering that each graph set has a set of distributions in terms of a graph property $x$, we first calculate the average distribution of the whole set. To get the average distribution of a graph set, the vectors of counts of the property $x$ of all the graphs are first concatenated. Then the probability densities of the graph property $x$ is calculated based on this concatenated vector as the average node degree distribution. Fianlly, the Kullback-Leibler divergence (KL-D~\cite{kullback1951information}) between the average node degree distribution of the generated graph set $P_\mathrm{ave}(x)$ and that of the real graph set $Q_\mathrm{ave}(x)$ is calculated as:\vspace{-0.2cm}

\vspace{-0.2cm}
\small
\begin{equation}\vspace{-0.1cm}\nonumber
    KL-D(P_\mathrm{ave},Q_\mathrm{ave})=-\sum\nolimits_{x\sim P_\mathrm{ave}}P_\mathrm{ave}(x) log(Q_\mathrm{ave}(x)/P_\mathrm{ave}(x)).
\end{equation}\normalsize

\emph{Maximum Mean Discrepancy (MMD)}~\cite{gretton2012kernel}. First, the squared MMD between the graph statistics distribution of the generated graph set $P$ and that of the real graph set $Q$ can be derived as:\vspace{-0.2cm}

\vspace{-0.2cm}
\small
\begin{align}\vspace{-0.1cm}\nonumber
    MMD(P,Q)=&\mathbb{E}_{x,y\sim P}[k(x,y)]+\mathbb{E}_{x,y\sim Q}[k(x,y)]\\&-2\mathbb{E}_{x\sim P,y\sim Q}[k(x,y)],
\end{align}\normalsize
where $x$, $y$ refer to the graph statistics that are sampled from the two distributions. The kernel $k(*)$ is designed as follows:\vspace{-0.2cm}
\begin{equation}\vspace{-0.1cm}
    k(x,y)=exp(W(x,y)/2\sigma^2),
\end{equation}\normalsize
where $\sigma$ refers to the standard deviation of $P$ or $Q$. Considering the sampled graph statistics are also two distributions, thus, $W(x,y)$ is defined as the Wasserstein distances (WD):\small
\begin{equation}
    W(x,y)=\mathop{inf}\limits\nolimits_{\gamma\in \prod(x,y)}\mathbb{E}_{(i,j)\sim \gamma}[\parallel i-j \parallel],
\end{equation}\normalsize
where $\prod(x,y)$ is the set of all measures whose marginals are $x$ and $y$ respectively.\\

\noindent\textbf{Distance metrics for scalar-valued statistics}.
The calculation of distance between two sets of graphs in terms of the scalar-valued statistic is much easier than that of distribution statistics. There are two major ways: (1) calculating the difference between the averaged value of the scalar-valued statistic of the generated graph set and that of the real graph set; (2) calculating the distance between the distribution of the scalar-valued statistic of the generated graph set and that of the real graph set. Many distance metrics can be used, such as KL-D, Jensen-Shannon distances (JS), and the Hellinger distance (HD).

\subsubsection{Classifier-based}
Classifier-based evaluation typically utilizes a graph classifier to evaluate whether the generated graphs follows the same distribution as the real graphs without explicitly defining the graph statistics. 
Typically, a classifier is trained on the set of real graphs and is tested on the set of generated graphs.
It only could be utilized when multiple graph generative models are trained for generating multiple types of graphs, respectively.
Here we introduce two existing classifier-based evaluations~\cite{liuauto} that are based on graph isomorphism network (GIN)~\cite{xu2019powerful} as follows.\\

\noindent\textbf{Accuracy-based}. First, a GIN is pre-trained on the training set consisting of multiple types of graphs previously used for training the generative model. Then for each type of generated graph, the classification accuracy of classifying this type of generated graphs based on the trained GIN is the final evaluation metric.\\

\noindent\textbf{Fréchet Inception Distance (FID)-based}. FID computes the distance in the embedding space between two multivariate Gaussian distributions fitted to a generated set and a test set. A lower FID value indicates better generation quality and diversity. For each type of graph, first, the generated and real graphs in the testing set are inputted into the pre-trained GIN to get the graph embeddings. Then the means $\mu_\mathrm{G}$ and
covariance matrices $\sum_\mathrm{G}$ of the embeddings of the generated graph set, and the means $\mu_\mathrm{R}$ and covariance matrices $\sum_\mathrm{R}$ of real graphs are estimated. Finally, the FID metric for this type of graphs is computed as follows:\vspace{-0.2cm} 

\vspace{-0.2cm}
\small
\begin{equation}\nonumber
    FID=\parallel \mu_\mathrm{G} -\mu_\mathrm{R} \parallel^2_2+\mathrm{Tr}(\sum\nolimits_\mathrm{G}+\sum\nolimits_\mathrm{R}-2(\sum\nolimits_\mathrm{G}\sum\nolimits_\mathrm{R})^{\frac{1}{2}}),
\end{equation}\normalsize
where $\mathrm{Tr}(\cdot)$ refers to the trance of a matrix.

\subsubsection{Intrinsic-quality-based}
Besides the evaluation by measuring the similarity between the real and generated graphs, there are three additional metrics that directly evaluate the quality of the generated graphs: their validity, uniqueness and novelty.\\

\noindent\textbf{Validity}. Since sometimes the generated graphs are required to preserve some properties, it is straightforward to evaluate them by judging whether they satisfy such requirements, such as the following:
(1) Cycles graphs/Tree graphs: Cycles and trees are graphs that have obvious structural properties and the validity is calculated as what percentage of generated graphs are actually cycles or trees~\cite{li2018learning}.
(2) Molecule graphs: Validity for molecule generation is the percentage of chemically valid molecules based on some domain specific rules~\cite{popova2019molecularrnn}.\\

\noindent\textbf{Uniqueness}.
Ideally, high-quality generated graphs should be diverse and similar, but not identical. Thus, uniqueness is utilized to capture the diversity of generated graphs~\cite{bacciu2020edge,goyal2020graphgen,madhawa2019graphnvp,li2018learning,popova2019molecularrnn}. To calculate the uniqueness of a generated graph, the generated graphs that are sub-graph isomorphic to some other generated graphs are first removed. The percentage of graphs remaining after this operation is defined as Uniqueness. For example, if the model generates 100 graphs, all of which are identical, the uniqueness is 1/100 = 1\%. \\

\noindent\textbf{Novelty}. Novelty measures the percentage of generated graphs that are not sub-graphs of the training graphs and vice versa~\cite{bacciu2020edge,goyal2020graphgen,madhawa2019graphnvp}. Note that identical graphs are defined as graphs that are sub-graph isomorphic to each other. In other words, novelty checks if the model has learned to generalize unseen graphs.

\subsection{Evaluation for conditional deep graph generation}
In addition to the above general evaluation metrics for graph generation, for conditional deep generative models for graph generation, some additional evaluation metrics can be involved, including: graph-property-based and mapping-relationship-based evaluations.

\subsubsection{Graph-property-based}
Considering that each of generated graph can have its associated real graph as label in the conditional graph generation task, we can directly compare each generated graph to its label graph by measuring their similarity or distance based on some graph properties or kernels, such as the following: (1) random-walk kernel similarity by using the random-walk based graph kernel~\cite{kang2012fast}; (2) combination of Hamming and Ipsen-Mikhailov distances(HIM)~\cite{jurman2015him}; (3) spectral entropies of the density matrices; (4) eigenvector centrality distance~\cite{bonacich1987power}; (5) closeness centrality distance~\cite{freeman1978centrality}; (6) Weisfeiler Lehman kernel similarity~\cite{shervashidze2011weisfeiler};
(7) Neighborhood Sub-graph Pairwise Distance Kernel~\cite{goyal2020graphgen} by matching pairs of sub-graphs with different radii and distances. 

\subsubsection{Mapping-relationship-based}
Mapping-relationship-based evaluation measures whether the learned relationship between the conditions and the generated graphs is consistent with the true relationship between the conditions and the real graphs.\\

\noindent\textbf{Explicit mapping relationship}.
In the situation where the true relationship between the input conditions and the generated graphs is known in advance, the evaluation can be conducted as follows: 
(1) When the condition is a category label, we can examine whether the generated graph falls into the conditional category by utilizing a graph classifier~\cite{sun2019graph,fan2019labeled}. Specifically, the real graphs are used to train a classifier and the classifier is used to classify the generated graphs. Then the accuracy is calculated as the percentage of the predicted categories that are the same as the input condition.
(2) When the condition is a graph, where the task is to change some properties of the input graph, we can quantitatively compare the property scores of the generated and input graphs to see if the change indeed meets the requirement. For example, one can compute the improvement of \textit{logP} scores of the optimized molecule in molecule optimization task~\cite{you2018graph}.\\

\noindent\textbf{Implicit mapping relationship}.
Regarding the deep graph translation problem, which is introduced
in Section~\ref{sec:graph translation}, sometimes, the underlying patterns of the mapping from the input graphs to the real target graphs are implicit and complex to define and measure. Thus, a classifier-based evaluation metric can be utilized~\cite{guo2022deep}. By regarding the input and target graphs as two classes, it assumes that a classifier that is capable of distinguishing the generated target graphs would also succeed in distinguishing the real target graphs from the input graphs. Specifically, a graph classifier is first trained based on the input and generated target graphs. Then this trained graph classifier is tested to classify the input graph and real target graphs, and the results will be used as the evaluation metrics.

\section{Applications}
\label{sec:six}
Deep generative models for graph generation is a very active research domain with a continuously increasing number of applications being proposed, including important topics such as molecule optimization and generation, semantic parsing in NLP, code modeling, and pseudo-industrial SAT instance generation.

\subsection{Molecule generation}
Molecule generation is a challenging mathematical and computational problem in drug discovery and material science; its aim is to design novel molecules under a range of chemical properties. Any small perturbation in the chemical structure may result in a large variation in the desired molecular property. Besides, the space of valid molecules quickly becomes prohibitively huge and complex as the number of combinatorial permutations of atoms and bonds grows. Currently, most drugs are hand-crafted by human experts in chemistry and pharmacology. The recent advances of deep generative models for graph generation has opened a new research direction by treating the molecule as a graph with atoms as nodes and bonds as edges, with the potential to learn these molecular' generative representation for novel molecule generation to ensure chemical validity and efficiency~\cite{jin2018junction,popova2019molecularrnn,zhang2019d,de2018molgan,samanta2019nevae,lim2019scaffold}. 

\begin{figure}[htb]\vspace{-0.3cm}
    \centering
    \includegraphics[width=0.27\textwidth]{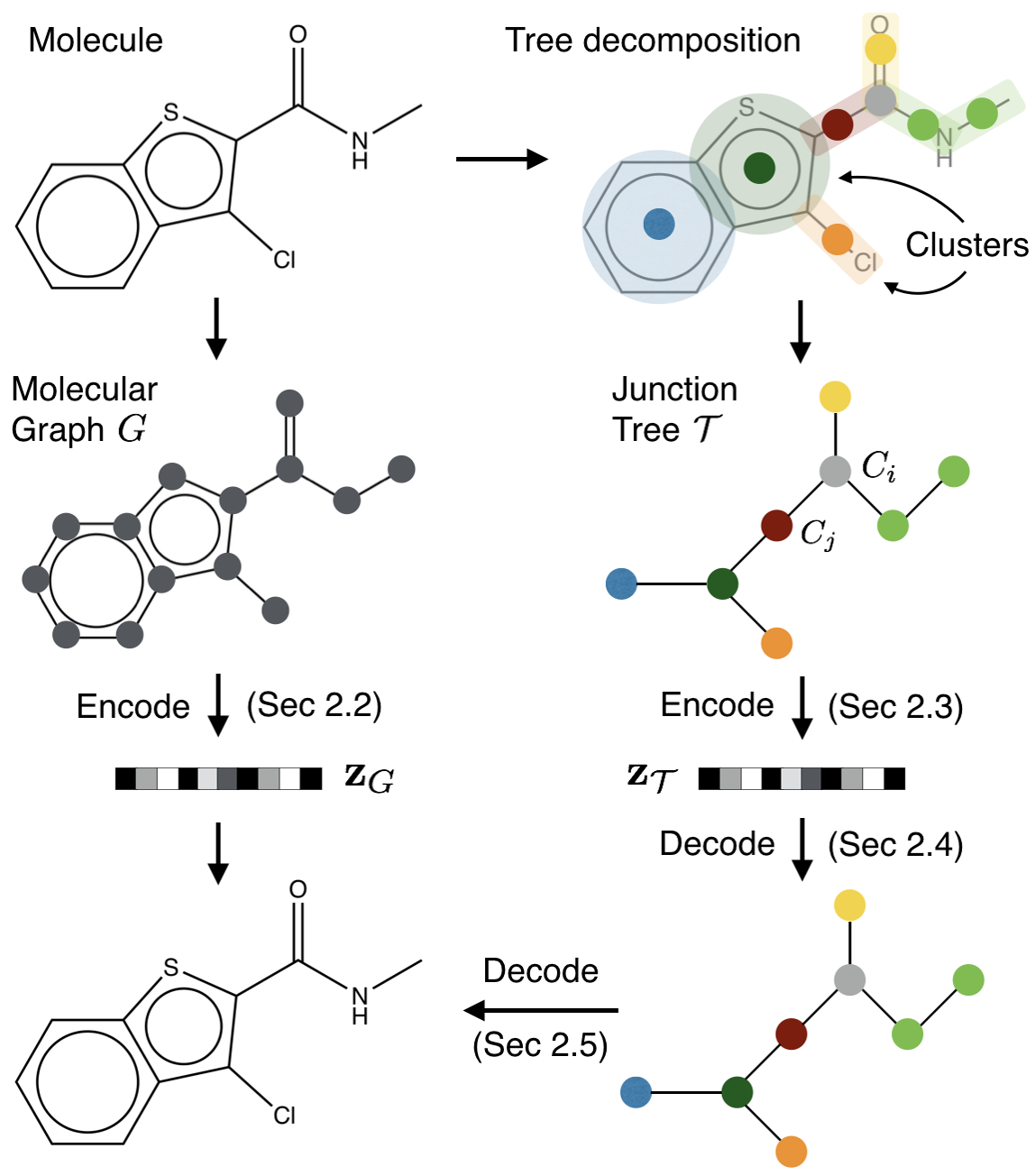}\vspace{-0.2cm}
    \caption{Framework overview of JT-VAE\cite{jin2018junction}: A molecular graph $G$ is first decomposed into its junction tree $\mathcal{T}_G$, where each colored node in the tree represents a substructure in the molecule. Then both the tree and graph are encoded into their latent embeddings $z_T$ and $z_G$. To decode, the junction tree is first reconstructed from $z_T$.}
    \label{fig:app_molecule}\vspace{-0.3cm}
\end{figure}

\noindent\textbf{Representative Work}. Junction Tree VAE (JT-VAE)~\cite{jin2018junction} formalizes the molecular structures generation task into an unconditional graph generation problem, where each atom in a molecule is a node in the graph and the bonds between atoms are represented as edges. JT-VAE adopts a motif-sequence-based generation approach, one of a number of sequential-based generating techniques, to generates a molecular graph by sequentially expanding a generated molecule by adding a valid chemical substructure in each step. Figure~\ref{fig:app_molecule} shows a backbone VAE-based generative model consisting of two encoders and decoders. Here, the molecular graph $G$ is first decomposed into its junction tree $\mathcal{T}_G$, where each colored node in the tree represents a substructure in the molecule. Then both the tree and graph are encoded into their latent embeddings $z_T$ and $z_G$. To decode the molecule, first step is to  reconstruct the junction tree from $z_T$ , and then assemble nodes in the tree to return to the original molecule. 




\subsection{Protein structure modeling}
Proteins are massive molecules that can be characterized as one of the multiple long chains of amino acids. Analyzing the structure and function of proteins is a key part of understanding biological properties at the molecular and cellular level. Current computational modeling methods for protein design are slow and often require human oversight and intervention, which are often biased and incomplete. Inspired by recent momentum in deep graph generative models, some works~\cite{guo2020generating,anand2018generative,golkov2016protein,livi2016generative,du2022interpretable} demonstrate the potential of deep graph generative modeling for fast generation of new, viable protein structures.\\

\begin{figure}[htb]\vspace{-0.3cm}
    \centering
    \includegraphics[width=0.38\textwidth]{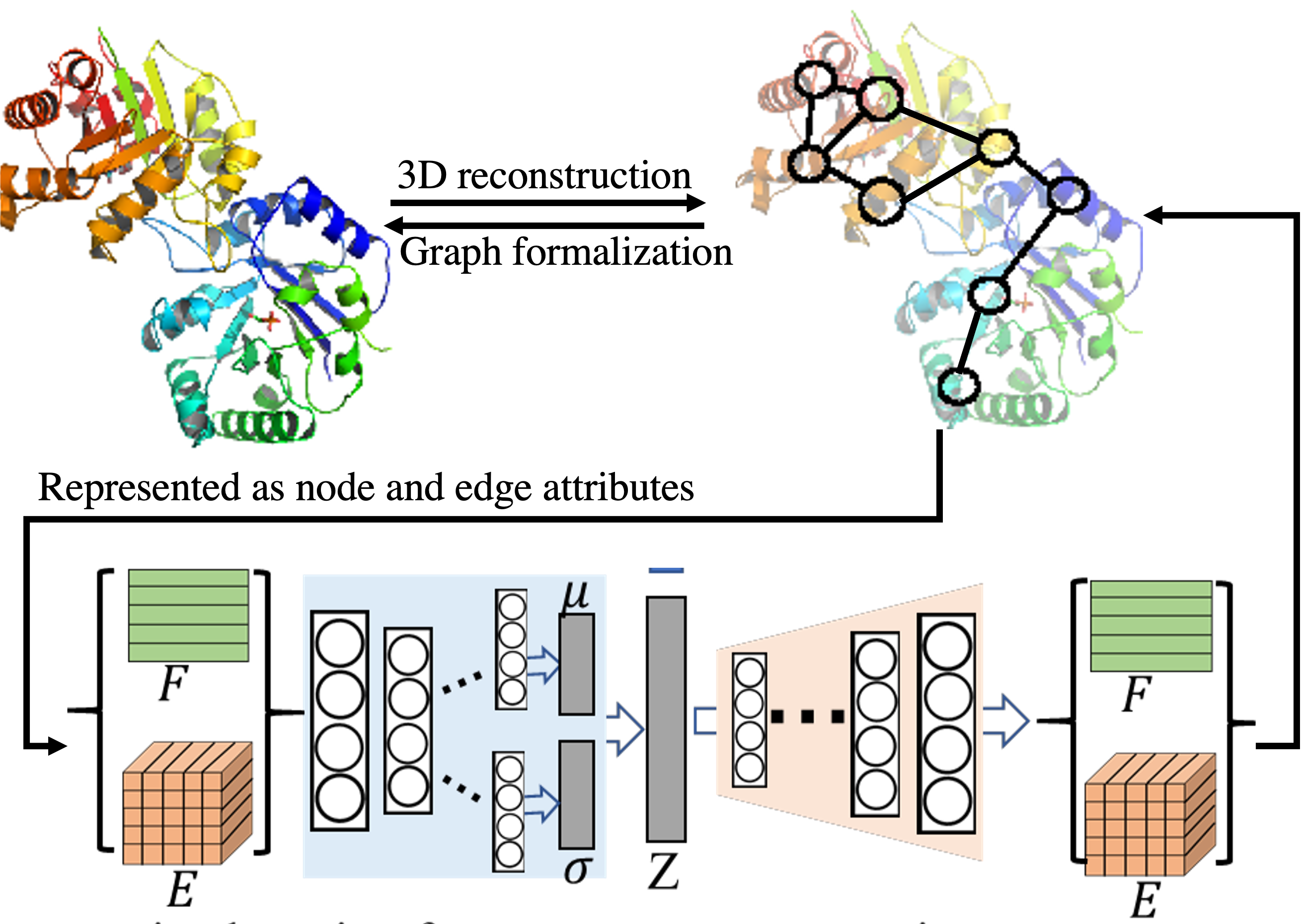}\vspace{-0.2cm}
    \caption{CO-VAE for protein structure modeling\cite{guo2020generating}: a protein graph represents the mutual distance between each pair of amino acids. Node and edge attributes are input into the encoder to learn the distribution of the latent embedding.}
    \label{fig:app_protein}\vspace{-0.2cm}
\end{figure}

\noindent\textbf{Representative Work}.
Guo et al~\cite{guo2020generating} proposed a contact VAE (CO-VAE) to generate functionally relevant three-dimensional protein structures. Here, the protein structure is formalized as a graph where each amino acid is a node and the physical distance between two amino acids determines the existence of an edge based on a pre-defined threshold. A graph generative model VAE is utilized to model and generate the graph by following the adjacent-matrix-based one-shot generating technique, where the node attributes and adjacent matrix of graph are generated in a single shot. As shown in Figure~\ref{fig:app_protein}, a protein structure is first represented by a graph that consists of a node attribute matrix and edge attribute tensor. These two components are then input into the encoder of VAE to learn the distribution of the latent embedding of the graph. In the decoder, the node and edge attributes are generated based on the sampled latent embedding and can then be recovered to yield the protein based on a 3D reconstruction technique.   

\vspace{-0.2cm}
\subsection{Semantic parsing}
Semantic parsing problem is about mapping the natural language information to its logical forms, namely abstract meaning representation (AMR).
Traditional semantic parsers are usually based on compositionally and manually designed grammar to create the structure of AMR, and used lexicons for semantic grounding, which is time-consuming and heuristic.
Recent works develops neural semantic parser with sequence-to-sequence models~\cite{dong2016language,jia2016data}, which, however, only consider the word sequence information and ignore other rich syntactic information. Because AMR are naturally structured objects (e.g. tree structures), semantic AMR parsing methods based on deep graph generative models are deemed as promising~\cite{wang2018neural,chen2018sequence,fancellu2019semantic,lyu2018amr,zhang2019amr}. These methods represent the semantics of a sentence as a semantic graph (i.e., a sub-graph of a knowledge base) and treat semantic parsing as a semantic graph matching/generation process. \\

\begin{figure}[htb]\vspace{-0.3cm}
    \centering
    \includegraphics[width=0.27\textwidth]{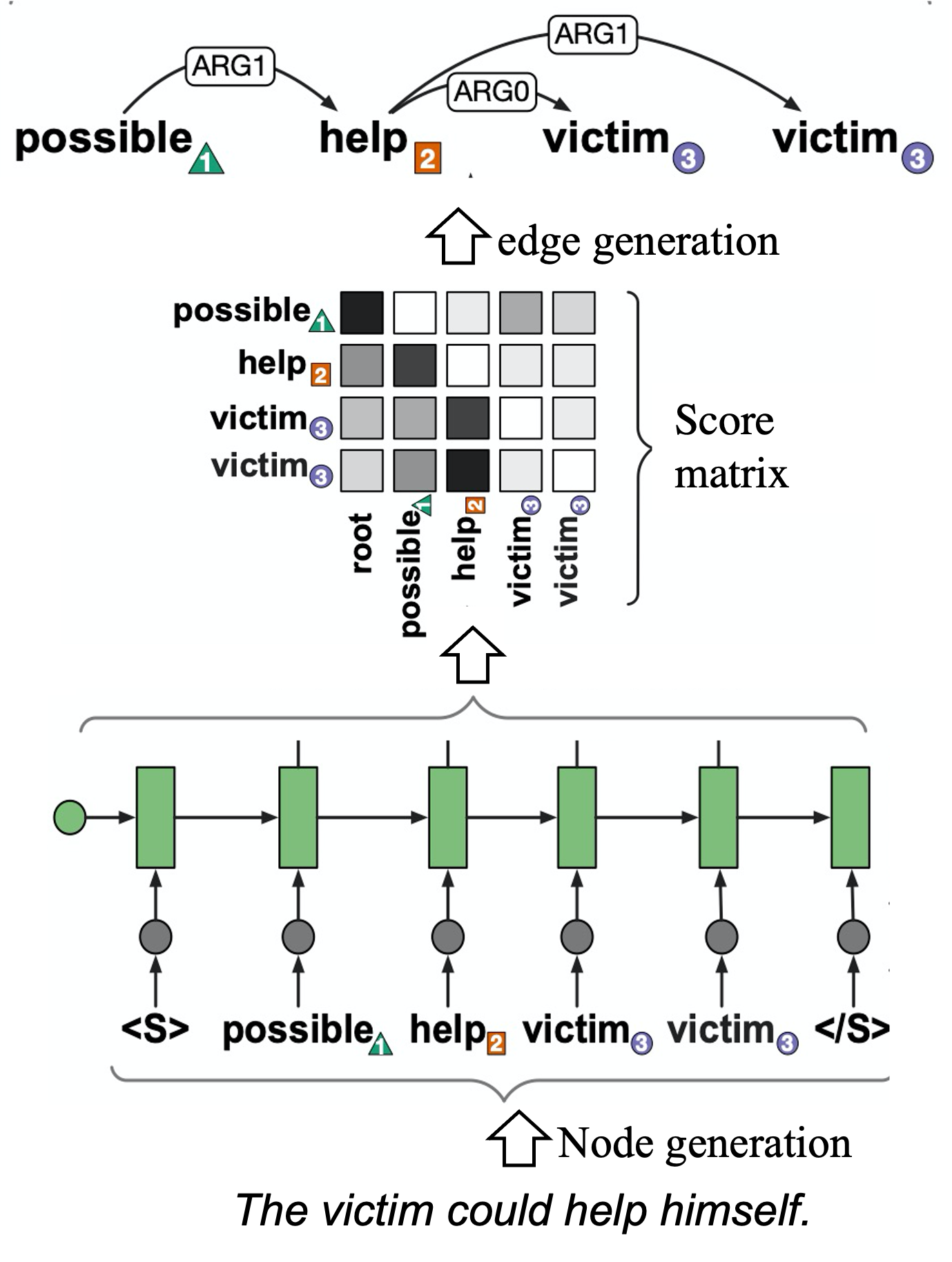}\vspace{-0.2cm}
    \caption{A two-stage AMR parsing process for a sequence-to-graph problem~\cite{zhang2019amr}: node prediction is to generate nodes based on the input sequence of tokens and edge generation generates the edges by sampling from the score matrix calculated based on the node representations.}
    \label{fig:app_amr}\vspace{-0.2cm}
\end{figure}

\noindent\textbf{Representative Work}.
Zhang et al\cite{zhang2019amr} formalized the AMR parsing as a graph generation problem conditioned on sequence, where the input is the sequence of tokens from a target sentence and the output its AMR graph. In the AMR graph, a node denotes to a word in the sentence and a predicted edge represents the semantic relationship between two words. This work is an edge-list-based one shot generation method, where the edges are generated based on pairs of node representations. As shown in Figure~\ref{fig:app_amr}, the whole process consists of two stages: node and edge prediction. The node prediction utilizes an RNN-based generative models to generate the nodes selected from the tokens in the sentence. For the second stage (i.e., the edge prediction), a score matrix that measures the probability of the edge existence is learnt based on the representation vectors of each pair of nodes, after which the edge is generated by sampling from the score matrix. This end-to-end deep graph generation techniques for semantic parsing has demonstrated a powerful ability for automatically capturing semantic information.

\vspace{-0.3cm}
\subsection{Code modeling}
Code modelling considers both hard syntactic and semantic constraints in generating natural programming code, which can make the development of source code easier, faster, and less error-prone. Early works in this area have shown that approaches from natural language processing can be applied successfully to the source code. However, though these methods are successful at generating programs that satisfy some formal specifications, they cannot generate realistic-looking and valid programs. Since \emph{program graphs} have been shown to have the ability to encode semantically meaningful representations of programs, deep graph generative models have shown promising capability in modeling small but semantic programs generation~\cite{brockschmidt2019generative,dai2018syntax,maddison2014structured,nguyen2015graph}. \\

\begin{figure}[htb]\vspace{-0.2cm}
    \centering
    \includegraphics[width=0.3\textwidth]{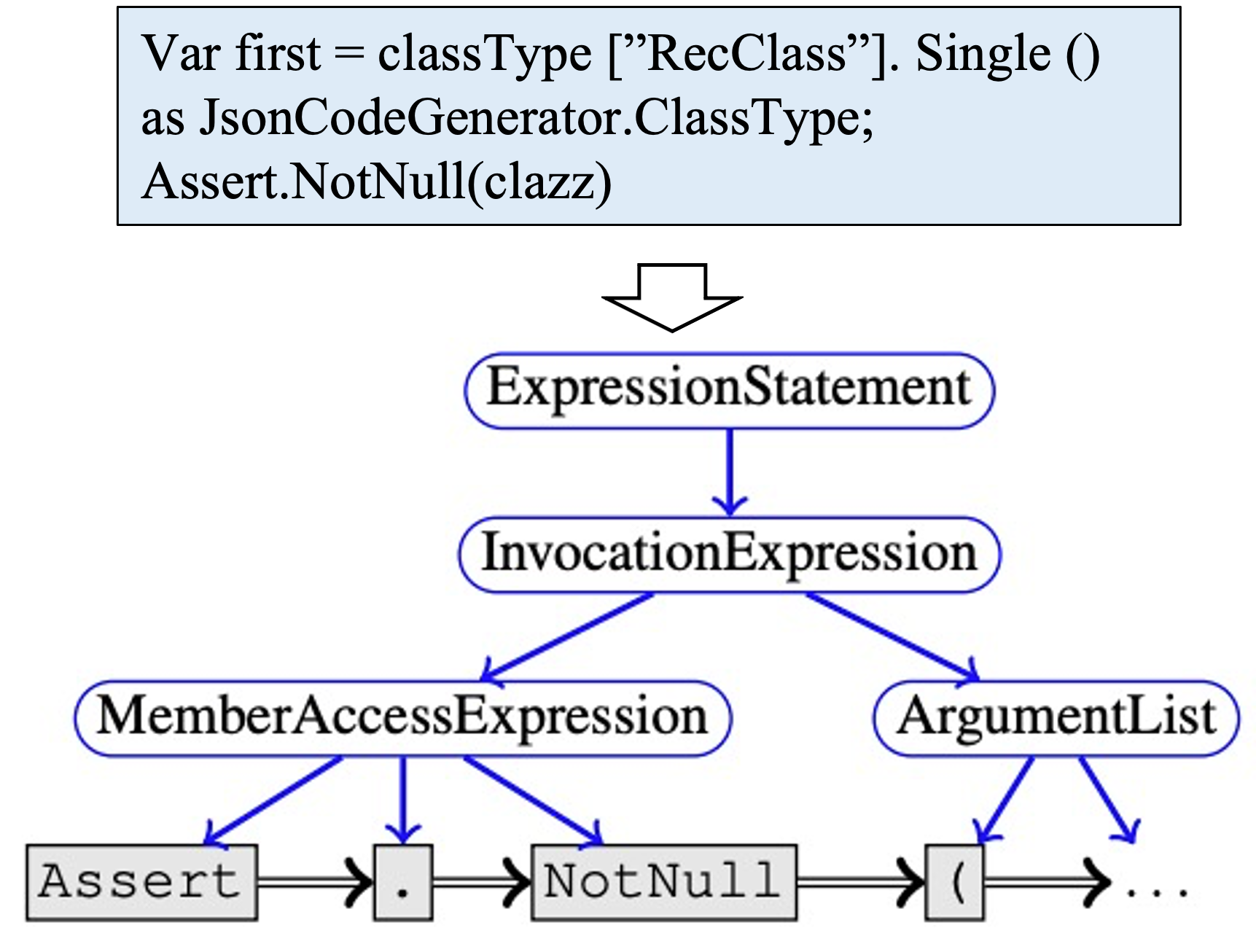}\vspace{-0.2cm}
    \caption{Representing a program as an abstract syntax tree \cite{brockschmidt2019generative,allamanis2018learning}: each node refers to a construct occurring in the codes and the edges denote to the semantic relationships.}
    \label{fig:app_code}\vspace{-0.2cm}
\end{figure}

\noindent\textbf{Representative Work}.
Brockschmidt et al.~\cite{brockschmidt2019generative} formalized the code modeling as a graph structure generation problem, where the source code is represented by an abstract syntax tree, as shown in Figure~\ref{fig:app_code}. In this abstract syntax tree, each node refers to a construct occurring in the codes and the edges denote to the semantic relationships. The generation process is a rule-based sequentially generating techniques, where the code is represented as an abstract syntax tree (AST), which incorporates rich structural information. An AST is then generated by expanding one node at a time using production rules from the underlying programming language grammar. This simplifies the code generation task to a sequence of sampling problems, in which an appropriate production rule must be sampled based on the partial AST generated so far.

\subsection{Pseudo-industrial SAT instance
generation}
The problem of pseudo-industrial Boolean Satisfiability (SAT) instance
generation is about generating artificial SAT problems that display the same characteristics as their real-world counterparts. Generating large amounts of SAT instances is important in developing and evaluating practical SAT solvers, which historically relies on extensive empirical testing on a large amount of SAT instances.
Prior works addressing this problem relied on hand-crafted algorithms, but have difficult in simultaneously
capturing a wide range of characteristics exhibited by real-world SAT instances~\cite{newsham2014impact,giraldez2015modularity}.  
Thus, it is promising to represent SAT formulas as graphs, thus recasting the original problem as a deep graph generation task~\cite{you2019g2sat, wu2019learning}. \\

\begin{figure}[htb]\vspace{-0.2cm}
    \centering
    \includegraphics[width=0.46\textwidth]{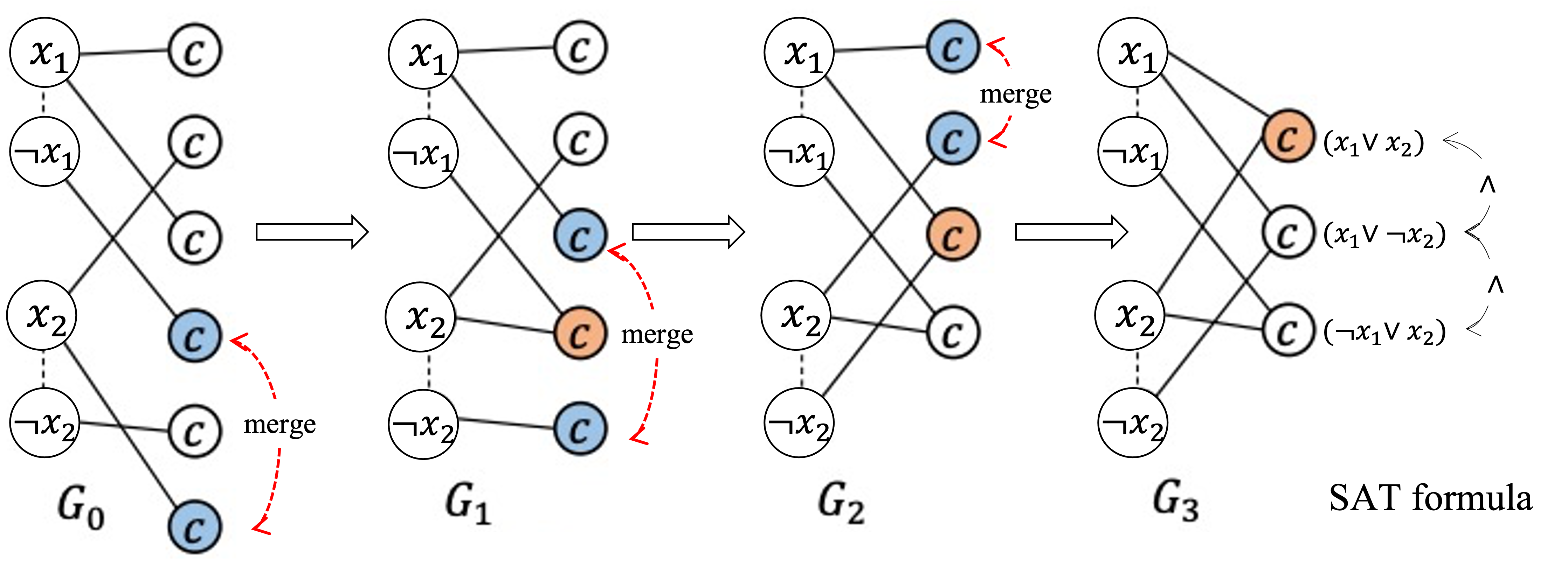}\vspace{-0.2cm}
    \caption{An overview of the G2SAT model~\cite{you2019g2sat}: In each step, two clause nodes are merged into a single clause node. A GCN-based classifier that captures the bipartite graph structure is used to sequentially decide which nodes to merge.}
    \label{fig:app_sat}\vspace{-0.2cm}
\end{figure}

\noindent\textbf{Representative Work}.
G2SAT~\cite{you2019g2sat} formalizes the SAT generation task as a graph generation problem by representing the SAT as a bipartite graph, where each node represents either a literal or a clause, with an edge denoting the occurrence of a literal in a clause representing a disjunction operation. In  general, the generation process is a motif-sequence-based generating style where a new motif is added to the partially generated graph in each step. The motifs refer to the trees that are split from the existing training bipartite graphs. As shown in Figure~\ref{fig:app_sat}, while generating, G2SAT generates a bipartite graph by starting with a set of motifs. In each step, a new motif is added by merging one of its clause nodes with an existing node in the partially generated graph. At last, all the conjunction clauses are combined with conjunction operations to recover the SAT formula. 

\section{Future Opportunities}
\label{sec:seven}
As a fast-developing, promising domain, there are still many open challenges in the domain of deep generative models for graph generation. In this section, we highlight a number of open challenges for further research.\\

\noindent\textbf{Scalability}.
Existing deep generative models typically have super-linear time complexity to the number of nodes and cannot scale well to large networks. Only few methods have linear time complexity of $O(N)$~\cite{shi2020graphaf,goyal2020graphgen,you2018graphrnn,bacciu2019graph,liuauto} and $O(M)$~\cite{samanta2019nevae}, where $N$ is the number of nodes and $M$ is the number of edges. Consequentially, most existing works merely focus on small graphs, typically with dozens to thousands of nodes~\cite{simonovsky2018graphvae,li2018learning,you2018graph,de2018molgan,grover2019graphite,salha2019gravity,fan2019labeled}. However, many real-world networks are large, with millions to billions of nodes ~\cite{goyal2020graphgen}, such as the Internet, biological neuronal networks, and social networks. It is important for any generative model to scale to large graph.\\

\noindent\textbf{Validity constraint}.
Many real-world networks are constrained by specific validity requirements~\cite{ma2018constrained}. For example, in molecular graphs, the number of bonding-electron pairs cannot exceed the valency of an atom. In protein interaction networks, two proteins may be connected only when they belong to the same or correlated gene ontology terms. 
Graph-topological constraints are challenging to enforce during the model training process. Intuitive ways include designing heuristic and customized algorithms to ensure the validity of generated graphs. For example, 
Dai et al.~\cite{dai2018syntax} further apply attribute grammar as a constraint in the parse-tree generation, a step toward semantic validity. 
Some recent works started to construct a more generic framework  under constrained optimization scenario, which minimizes training loss under graph validity constraints~\cite{ma2018constrained}. However, as such constraints are typically discrete and non-differentiable, they need to be approximated with a smooth relaxation which introduces errors and cannot preclude all the invalid topologies.\\

\noindent\textbf{Interpretability}.
When we learn the underlying distribution of complex structured data, i.e. graphs, learning interpretable representations of data that expose semantic meaning is very important~\cite{lake2017building}. For example, it is highly beneficial if we could identify which latent variable(s) control(s) which specific properties (e.g., molecule mass) of the generated graphs (e.g., molecules). It is also useful to disentangle local generative dependencies among different sub-graphs. However, existing works on this topic only focus on graph embedding but not generation~\cite{noutahi2019towards,bouchacourt2018multi}. For example, Stoehr et al~\cite{stoehr2019disentangling} demonstrates the potential of latent variable disentanglement in graph deep learning for unsupervised discovery of generative parameters of random and real-world graphs. Investigations on graph decoding and generation are still open problems without existing works except very recently published ones~\cite{guo2020interpretable,du2022disentangled, guo2021deep}.  \\

\noindent\textbf{Beyond training data}.
Deep generative models are data-driven models based on training data. The novelty of the generated graphs are highly desired yet usually restricted by training data and model properties (e.g., mode collapse of generative adversarial nets). To address such issues, attempts in the domain of images modified the attribute of a generated image by adding a learned vector on its latent code~\cite{radford2016unsupervised} or by combining the latent code of two images~\cite{karras2019style}. Additional works have been developed for inserting extra control in the image generation ~\cite{radford2016unsupervised} with additional labels corresponding to key factors such as object size and facial expression. However, works on graph generation that could require very different technique sets than image generation are lacking. \\

\noindent\textbf{Dynamic graphs}.
Existing deep graph generative models typically focus on static graphs but many graphs in the real-world are dynamic, and their node attributes and topology can evolve over time, such as social network, mobility network, and protein folding. Representation learning for dynamic graphs is a hot domain, but it only focuses on graph embedding instead of generation. Modeling and understanding the generation of dynamic graphs have not been explored. Therefore, additional problems such as jointly modeling temporal and graph patterns and temporal validity constraints need to be addressed.

\section{Conclusion}
\label{sec:eight}
In this survey paper, we provides a systematic review of deep generative models for graph generation. We present a taxonomy of deep graph generative models based on problem settings and techniques details, followed by a detailed introduction, comparison, and discussion about them.
We also conduct a systematic review of the evaluation measures of deep graph generative models, including the general evaluation metrics for both unconditional and conditional graph generation. After that, we summarized popular applications in this domain.

\bibliographystyle{IEEEtran}
\bibliography{main}

\appendices
\section{Preliminaries Knowledge of Deep Generative Models}
\label{app1}
In recent years, there has been a resurgence of interest in deep generative models, which have been at the forefront of deep unsupervised learning for the last decade. The reason for that is because they offer a very efficient way to analyze and understand unlabeled data. The idea behind generative models is to capture the inner probabilistic distribution that generates a class of data to generate similar data~\cite{oussidi2018deep}. Emerging approaches such as generative adversarial networks (GANs)~\cite{goodfellow2014generative}, variational auto-encoders (VAEs)~\cite{kingma2014auto}, generative recursive neural network (generative RNN)~\cite{sutskever2011generating} (e.g., pixelRNNs, RNN language models), flow-based learning~\cite{papamakarios2017masked}, and many of their variants and extensions have led to impressive results in myriads of applications. 
In this section, we provide a review of five popular and classic deep generative models for learning the distributions by observing large amounts of data in any format. They include VAE, GANs, generative RNN, flow-based learning, and Reinforcement Learning, which also form the backbone of the base learning methods of all the existing deep generative models for graph generation.
\begin{figure}[htb]\vspace{-0.4cm}
    \centering
    \includegraphics[width=0.5\textwidth]{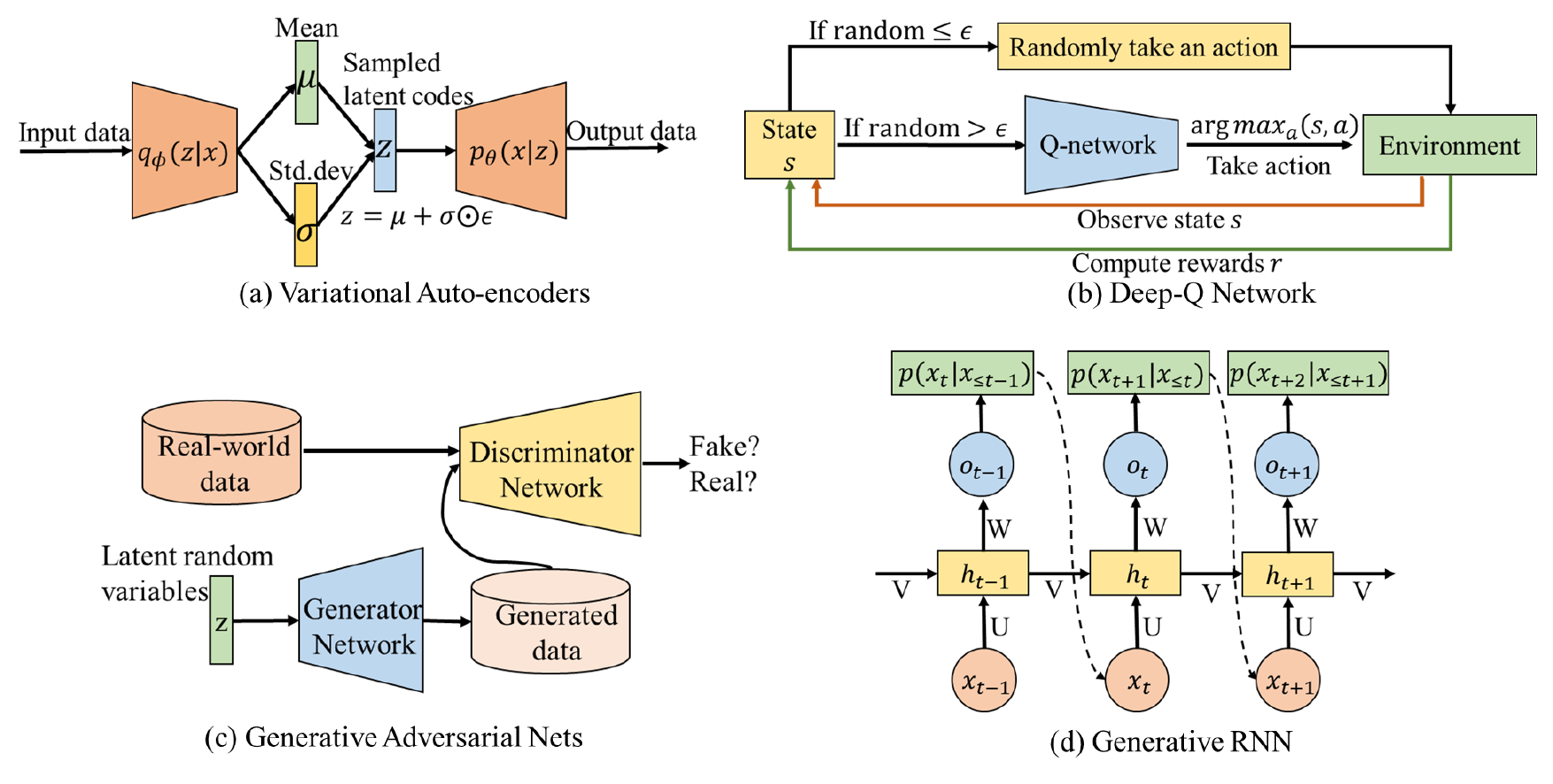}\vspace{-0.4cm}
    \caption{Abstract architecture of deep generative models: (a) Variational auto-encoders; (b) Deep Q-network; (c) Generative adversarial nets; (d) generative RNN.}\vspace{-0.4cm}
    \label{fig:deep-generative-model}
\end{figure}



\subsection{Variational Auto-encoders}
VAE~\cite{kingma2014auto} is a latent variable-based model that pairs a top-down generator with a bottom-up inference network. Instead of directly performing maximum likelihood estimation on the intractable marginal log-likelihood, training is done by optimizing the tractable evidence lower bound (ELBO). Suppose we have a dataset of samples $x$ from a distribution parameterized by ground truth generative latent codes $z\in \mathbb{R}^c$ ($c$ refers to the length of the latent codes). VAE aims to learn a joint distribution between the latent space $z\sim p(z)$ and the input space $x\sim p(x)$.

Specifically, in the probabilistic setting of a VAE, the encoder is defined by a variational posterior
$q_\phi(z|x)$, while the decoder is defined by a generative distribution $p_\theta(x|z)$, as represented by the two orange trapezoids in Fig.~\ref{fig:deep-generative-model}(a).  $\phi,\theta$ are trainable parameters of the encoder and decoder. The VAE aims to learn a marginal likelihood of the data in a generative process as: $\underset{\phi,\theta}{max}\mathbb{E}_{q_\phi(z|x)}[\mathrm{log}p_\theta(x|z)]$.
Then the marginal likelihoods of individual data points can be rewritten as follows:
\begin{equation}
    \mathrm{log}p_\theta(x|z)=D_{KL}(q_\phi(z|x)||p(z))+\mathcal{L}(\phi,\theta; x,z),
\end{equation}
where the first term stands for the non-negative Kullback–Leibler divergence between the true and the
approximate posterior; the second term is called the (variational) lower bound on the marginal likelihood. Thus, maximizing $\mathcal{L}(\phi,\theta; x,z)$ is to maximize the lower bound of the true objective $
    \mathcal{L}(\phi,\theta; x,z)=\mathbb{E}_{q_\phi(z|x)}[\mathrm{log}p_\theta(x|z)]-D_{KL}(q_\phi(z|x)||p(z)).$
In order to make the optimization of the above objective tractable in practice, we assume a simple prior distribution $p(z)$ as a standard Gaussian $\mathcal{N}(\textbf{0}, \textbf{I})$ with a diagonal co-variance matrix. Parameterizing the distributions in this way allows for the use of the “reparameterization trick” to estimate gradients of the lower bound with respect to the parameter $\phi$, where each random variable $z_i\sim q_\phi(z_i|x)$ is parameterized as Gaussian with a differentiable transformation of a noise variable $\epsilon \sim \mathcal{N}(\textbf{0}, \textbf{1})$, that is, $z$ is computed as $z=\mu+\sigma\odot\epsilon$, where $\mu$ and $\sigma$ are outputs from the encoder.

\subsection{Generative Adversarial Nets}
GANs were introduced as an alternative way to train a generative model~\cite{goodfellow2014generative}. GANs are based on a game theory scenario called the min-max game, where a discriminator and a generator compete against each other. The generator generates data from stochastic noise, and the discriminator tries to tell whether it is real (coming from a training set) or fabricated (from the generator).
The absolute difference between carefully calculated rewards from both networks is minimized so that both
networks learn simultaneously as they try to outperform each other.

Specifically, the architecture of GANs consists of two ‘adversarial’ models: a generative model $\mathcal{G}_\theta$ which captures the data distribution $p(x)$, and a discriminative model $\mathcal{D}_\phi$ which estimates the probability that a sample comes from the training set rather than $\mathcal{G}_\theta$, as shown in Fig.\ref{fig:deep-generative-model}(c). Both $\mathcal{G}_\theta$ and $\mathcal{D}_\phi$ could be a non-linear mapping function, such as a multi-layer perceptron~\cite{suter1990multilayer} parameterized by parameters $\theta$ and $\phi$.
To learn a generator distribution $p_{model}(x)$ of observed data $x$, the generator builds a mapping function from a prior noise distribution $p_z(z)$ to data space as $\mathcal{G}_{\theta}(z)$. And the discriminator, $\mathcal{D}_{\phi}(x)$, outputs a single scalar representing the probability that the input data $x$ came form the training data rather than sampled from $p_{model}(x)$.

The generator and discriminator are both trained simultaneously by adjusting the parameters of $p_{model}(x)$ to minimize $log(1- \mathcal{D}_\phi(\mathcal{G}_\theta(z))$
and adjusting the parameters of $\mathcal{D}_\phi$ to minimize $log\mathcal{D}_\phi(x)$, as if they are following the two-player min-max game with value function $V(\mathcal{G}_\theta, \mathcal{D}_\phi)$:\vspace{-0.1cm}
\begin{align}\vspace{-0.15cm}\nonumber
    \underset{\mathcal{G}_\theta}{min}\underset{\mathcal{D}_\phi}{max}V(\mathcal{G}_\theta, \mathcal{D}_\phi)&=\mathbb{E}_{x\sim p_{model}(x)}[log\mathcal{D}_\phi(x)]\\&+\mathbb{E}_{z\sim p_z(z)}[log(1-\mathcal{D}_\phi(\mathcal{G}_\theta(z))],
\end{align}
The training of the generator and discriminator is kept alternating until the generator can hopefully generate real-like data that is difficult to discriminate from real samples by a strong discriminator. 

In general, GANs show great power in generating data such as image~\cite{denton2015deep,goodfellow2014generative}, audio~\cite{chen2017deep}, and texts~\cite{nam2018text}. In contrast to VAE, GANs
learn to generate samples without assuming an approximate distribution. By utilizing the discriminator, GANs avoid optimizing the explicit likelihood loss function, which may explain their ability to produce high-quality objects as demonstrated by~\cite{denton2015deep}.
However, GANs still have drawbacks. One is that they can sometimes be extremely hard to train in adversarial style. They may fall into the divergence trap very easily by getting stuck in a poor local minimum. Mode collapse is also an issue, where the generator produces samples that belong to a limited set of modes, which results in low diversity. Moreover, alternatively training and large computation workloads for two networks can result in long-term convergence process.

\subsection{Generative Recursive Neural Network}

RNN~\cite{mikolov2010recurrent} is a straightforward adaptation of the standard feed-forward neural network by using their internal state (memory) to process variable length sequential data. At each step, the RNN predicts the output depending on the previous computed hidden states and updates its current hidden state, that it, they have a “memory” that captures information about what has been calculated so far. The RNN’s high dimensional hidden state and nonlinear evolution endow it with great expressive power to integrate information over many iterative steps for accurate predictions. Even if the non-linearity used by each unit is quite simple, iterating it over time leads to very rich dynamics~\cite{sutskever2011generating}.

A standard RNN is formalized as follows: given a sequence of input vectors $(x_1,...,x_T )$, the RNN computes a sequence of hidden states $(h_1,..., h_T)$ and a sequence of outputs $(o_1,...,o_T)$ by iterating the following equations from $t=1$ to $T$:\vspace{-0.15cm}
\begin{align}\vspace{-0.15cm}
    &h_t=\mathrm{tanh}(Ux_t+Vh_{t-1}+b_h); \quad
    &o_t=Wh_t+b_o
\end{align}
where $U$, $V$, and $W$ are learning weight matrices; the vectors $b_h$ and $b_o$ are biases for calculating the hidden states and output at each step, respectively. The expression $Vh_{t-1}$ at step $t = 1$ is initialized by a vector, $h_0$, and the tanh non-linearity activation function is applied coordinate-wise.    

The RNN model can be modified to a generative model for generating the sequential data, as shown in Fig.~\ref{fig:deep-generative-model}(d). The goal of modeling a sequence is to predict the next element in the sequence given the previous generated elements. More formally, given a training sequence $(x_1,...,x_T)$, RNN uses the sequence of its output vectors $(o_1,...,o_T )$ to parameterize a sequence of predictive distributions $p(x_{t+1}|x_{\le t})$. The distribution type 
of $p(x_{t+1}|x_{\le t})$ need to be assumed in advance. For example, to determine the category of the discrete data $x_{t+1}$, we can assume a softmax distribution as $p(x_{t+1} = j) = exp(o^{(j)}_t)/\sum_K o_t^{(K)}$, where $j$ refers to one of the categories of the object, $o_t^{(j)}$ refers to the $j$-th variable in the output vector $o_t$ and $K$ refers to the total number of categories of the objects. The objective of modeling sequential data is to maximize the total log likelihood of the training sequence $\sum^{T-1}_{t=0}\mathrm{log}p(x_{t+1}|x_{\le t})$,
which implies that the RNN learns a joint probability distribution of sequences. Then we can generate a sequence by sampling from $p(x_{t+1}|x_{\le t})$ stochastically, which is parameterized by the output at each step.

\subsection{Flow-based Learning}
\label{sec:flow-based}
Normalizing flows (NFs)~\cite{dinh2016density} are a class of generative models that define a parameterized invertible deterministic transformation between two spaces $z$ and $x$. 
$z\sim p_z(z)$ is a latent space that follows distribution such as Gaussian, and $x\sim p_x(x)$ is a real-world
observational space of objects such as images, graphs, and texts. Let $f_\theta:z \xrightarrow{} x$ be an invertible transformation parameterized by $\theta$. Then the relationship between the density function of real-world data $x$ and that of $z$ can be expressed via the change-of-variables formula:\vspace{-0.2cm}
\begin{align}\vspace{-0.3cm}
    p_x(x)=p_{z}(f_{\theta}^{-1}(x))|\mathrm{det}(\partial f_{\theta}^{-1}(x)/\partial x)|.
\end{align}
There are two key processes of normalizing flows as a generative model: (1) Calculating data likelihood: given a datapoint $x$, the exact density $p_x(x)$ can be calculated by inverting the transformation $z=f_{\theta}^{-1}(x)$; (2) Sampling: $x$ can be sampled from the distribution $p_x(x)$ by first sampling $z \sim p_z(z)$ and then performing the transformation $x=f_\theta(z)$. To efficiently
perform the above mentioned operations, $f_\theta$ is required to be invertible with an easily computable Jacobian determinant. 

Autoregressive flow (AF), originally proposed in~\cite{papamakarios2017masked}, is a variant of normalizing flow by providing an easily computable triangular Jacobian determinant. It is specially designed for modeling the conditional distributions in the sequence. Formally, given $x\in \mathbb{R}^D$ (D is the dimension of observed sequential data), the autoregressive conditional probabilities for the $d$-th element in the sequence can be parameterized as Gaussian:\vspace{-0.1cm}
\begin{align}
    &p(x_d|x_{1:d-1})=\mathcal{N}(\mu_d,(\sigma_d)^2)
\end{align}
where $\mu_d=g_\theta(x_{1:d-1})$ and $\sigma_d=g_\phi(x_{1:d-1})$ ($g_\theta$ and $g_\phi$ are unconstrained and positive scalar functions of $x_{1:d-1}$ respectively for computing
the mean and deviation). In practice, these functions can be implemented as neural networks. The affine transformation of AF can be written as follows:\vspace{-0.2cm}
\begin{align}\nonumber
    f_\theta(z_d)=x_d=\mu_d+\sigma_d \cdot z_d;\quad
    f^{-1}_\theta(x_d)=z_d=(x_d-\mu_d)/\sigma_d,
\end{align}
where $z_d$ is the randomly sampled value from standard Gaussian. The Jacobian matrix here is triangular, since $\partial x_i/\partial z_j$ is non-zero only for $j\leqslant i$. Therefore, the determinant can be efficiently computed through
$\prod^D_{d=1}\sigma_d$. Specifically, to perform density estimation, we can apply all individual scalar affine transformations in parallel to compute the base density, each of which depends on previous variables $x_{1:d-1}$; to sample $x$, we can first sample $z\in \mathbb{R}^D$ and compute $x_1$ through the affine transformation, and then each subsequent $x_d$ can be computed sequentially based on $x_{1:d-1}$.

\subsection{Reinforcement Learning and Deep Q-Network}
\label{sec:RL}
Reinforcement learning (RL) is a commonly used framework for learning controlling policies by a computer algorithm, the so-called agent, through interacting with its environment~\cite{sutton1998introduction,silver2007reinforcement}. Here, we give a brief introduction of this learning strategy as well as its typical form deep Q-learning networks (DQNs)~\cite{mnih2015human} for data generation.

In RL process, an agent is faced with a sequential decision making problem, where interaction with the environment takes place at discrete time steps. The agent takes action $a_t$ at state $s_t$ at time $t$, by following certain policies or rules, which will result in a new state $s_{t+1}$ as well as a reward $r_t$. If we consider infinite horizon problems with a discounted cumulative reward objective $R_t=\sum_{t'=t}^{\infty}\gamma^{t'-t}r_{t'}$ ($\gamma \in [0,1]$ is the discount factor),
the aim of the agent is to find an optimal
policy $\pi: s\xrightarrow{} a$ by maximizing its expected discounted cumulative rewards.
Q-Learning~\cite{watkins1992q} is a value-based method for solving RL problems by encoding policies through the use of action-value functions:\vspace{-0.2cm}
\begin{equation}
    Q^{\pi}(s,a)=\mathbb{E}_{\pi}[\sum\nolimits_{t=0}^{\infty}\gamma^t r_t|s_0=s,a_0=a].
\end{equation}
The optimal value function is denoted as $Q^*(s, a)=\underset{\pi}{max}Q^{\pi}(s,a)$, and an optimal policy $\pi^*$ can be easily derived by $\pi^{*}(s)\in \mathrm{argmax}_a Q^{*}(s,a)$.
Typically, Q-value function relies on all possible state-action pairs, which are often impractical to obtain. One solution for addressing this challenge is to approximate $Q(s, a)$ using a parameterized function~\cite{sutton1998introduction}.

Based on recent advances in deep learning techniques, Mnih et al.~\cite{mnih2015human} introduced the DQN. The DQN approximates the Q-value function with a non-linear deep convolutional network, which also automatically creates useful features to represent the internal states of the RL, as shown in Fig.~\ref{fig:deep-generative-model}(b). In DQN, the agent interacts with the environment in $i$ discrete iterations, aiming to maximize its long term reward. DQN has shown great power in generating sequential objects by taking a series of actions~\cite{li2016deep}. A sequential object is generated based on a sequence of actions that are taken.

During the generation, DQN selects the action at each step using an $\epsilon$-greedy implementation. With probability $\epsilon$, a random action is selected from the range
of possible actions, otherwise the action which results in high Q-value score is selected.
To perform experience replay, the agent’s experiences $e_t= (s_t,a_t,r_t,s_{t+1})$ at each time-step $t$ are stored in a data set $D_t=\{e_1,…,e_t\}$. At each iteration $i$ in the learning process, the updates of the learning weights are applied on samples of experience $(s_t,a_t,r_t,s_{t+1})\sim U(D)$, drawn randomly from the pool of stored samples, with the following loss function:\vspace{-0.15cm}
\begin{align}\vspace{-0.15cm}\nonumber
    \mathcal{L}(\theta_i)=&\mathbb{E}_{(s_t,a_t,r_t,s_{t+1})\sim U(D)}[(r_t+\gamma \underset{a_{t+1}}{max} Q(s_{t+1},a_{t+1};\theta^{-}_i)\\&-Q(s,a;\theta_i))^2],
\end{align}
where $\theta_i$ refers to the parameters of the Q-network at iteration $i$ and $\theta^{-}_i$ refers to the network parameters used to compute the target at iteration $i$. The target network parameters $\theta^{-}_i$ are only updated with the Q-network parameters $\theta_i$ every several steps and are held fixed between individual updates. The process of generating the data after training is similar to that of the training process.

\section{Benchmark Results and Datasets}
\label{app2}

As deep graph generation is a relatively new research area, it is important to quantitatively compare the performance of the existing algorithms, and provide the unified benchmark dataset for the future new algorithms and research. In this section, we first summarize the existing benchmark datasets that are used to evaluate the existing models. Next, we compared the published results of the existing deep generative models by using the evaluation metrics introduced in Section 5~\footnote{We only consider unconditional graph generation in this section due to the small number of the existing conditional graph generation methods.}.

\begin{table*}[htb]
    \centering
    \caption{Quantitative evaluation and comparison on general graph generation tasks by different deep generative models on graphs ($N$ refers to the number of nodes in the graph, ``D." refers to MMD for node degree, ``C." refers to MMD for clustering coefficient distributions, and ``O." refers to MMD for average orbit counts statistics. ``-'' denotes the unavailability of the published results from the original papers).}
    \begin{tabular}{|c|ccc|ccc|ccc|c|c|}
    \toprule\hline
        \multirow{2}{*}{Method} & \multicolumn{3}{|l|}{Community}&\multicolumn{3}{|l|}{Ego}&\multicolumn{3}{|l|}{Protein}&\multirow{2}{*}{Complexity}&\multirow{2}{*}{Type}\\\cline{2-10}
        ~&D.&O.&C.&D.&O.&C.&D.&O.&C.&~&~\\\hline
        GraphGMG\cite{li2018learning}&0.220&0.950&0.400&0.040&0.100&0.020&-&-&-&$O(N!)$&Sequential\\\hline
        GraphVRNN\cite{su2019graph}&\textbf{0.015}&0.057&\textbf{0.005}&0.052&0.184&0.010&-&-&-&$O(N^2)$&Sequential\\\hline
        EDP-GNN~\cite{niu2020permutation}&0.053& 0.144& 0.026& 0.052& 0.093& 0.007&-&-&-&$O(N^2)$&One-shot\\\hline
        GNF~\cite{liu2019graph} & 0.200& 0.200& 0.110& 0.030& 0.100& \textbf{0.001}&-&-&-&$O(N^2)$&One-shot\\\hline
       GraphAF~\cite{shi2020graphaf} & 0.060& 0.100& 0.015& 0.040& \textbf{0.040}& 0.008&-&-&-&$O(N^2)$&One-shot\\\hline    
         GraphVAE\cite{simonovsky2018graphvae}&0.350&0.980&0.540&0.130&0.170&0.050&0.480&0.071&0.740&$O(N^4)$&One-shot\\\hline 
         GraphRNN\cite{you2018graphrnn}&0.030&\textbf{0.030}&0.010&\textbf{0.001}&0.050&\textbf{0.001}&0.034&0.935&0.217&$O(N^2)$&Sequential\\\hline
        GRAN\cite{liao2019efficient}&-&-&-&-&-&-&\textbf{0.002}&\textbf{0.048}&0.140&$O(N)$&Sequential\\\hline
        GRAN-I\cite{gu2019explore}&-&-&-&-&-&-&0.007&0.074&0.059&$O(N)$&Sequential\\\hline
        LGGAN~\cite{fan2019labeled}&-&-&-&-&-&-&0.180&0.150&\textbf{0.020}&$O(N^2)$&One-shot\\\hline\bottomrule   
    \end{tabular}
    \label{tab:comparison_general}
\end{table*}\normalsize

\begin{table*}[!tb]
    \centering
    \caption{Quantitative evaluation and comparison on molecule structure generation tasks by different deep generative models on graphs ($C$ refers to number of motifs and $N$ refers to number of nodes in the graph. ``Unique.'' is short for uniqueness. ``Novel.'' is short for novelty.``Valid.'' is short for validness).}
    \begin{tabular}{|c|rrr|rrr|l|c|}
    \toprule\hline
        \multirow{2}{*}{Method} & \multicolumn{3}{|l|}{ZINC}&\multicolumn{3}{|l|}{QM9}&\multirow{2}{*}{Complex.}&\multirow{2}{*}{Type}\\\cline{2-7}
        ~&Unique. &Novel. & Valid.&Unique. &Novel.&Valid.&~&~\\\hline
        GrammarVAE~\cite{kusner2017grammar}&10.76\%&\textbf{100.00\%}&31.00\%&9.30\%&95.44\%&30.00\%&-&Rule-sequential\\\hline
        GraphVAE\cite{simonovsky2018graphvae}&31.60\%&\textbf{100.00\%}&14.00\%&40.90\%&85.00\%&61.00\%&$O(N^4)$&One-shot\\\hline
        CGVAE\cite{liu2018constrained}&99.82\%&\textbf{100.00\%}&\textbf{100.00\%}&\textbf{98.54\%}&94.35\%&\textbf{100.00\%}&$O(N^2)$&Node-sequential\\\hline
        GraphNVP\cite{madhawa2019graphnvp}&94.8\%&\textbf{100.00\%}&74.30\%&97.30\%&54.00\%&90.10\%&$O(N^2)$&One-shot\\\hline
        GRF\cite{honda2019graph}&53.70\%&\textbf{100.00\%}&73.40\%&66.00\%&58.60\%&84.50\%&$O(N^2)$&One-shot\\\hline
        GraphAF\cite{shi2020graphaf}&99.10\%&\textbf{100.00\%}&\textbf{100.00\%}&94.51\%&88.83\%&\textbf{100.00\%}&$O(N^2)$&One-shot\\\hline        
        CGSVAE\cite{ma2018constrained}&-&\textbf{100.00\%}&34.90\%&-&\textbf{97.50\%}&96.60\%&$O(N^2)$&One-shot\\\hline
        JT-VAE~\cite{jin2018junction}&\textbf{100.00\%}&\textbf{100.00\%}&99.80\%&-&-&-&$O(C)$&Motif-sequential\\\hline
        GCPN~\cite{you2018graph}&99.97\%&\textbf{100.00\%}&\textbf{100.00\%}&-&-&-&$O(C)$&Motif-sequential\\\hline
        MolecularRNN~\cite{popova2019molecularrnn}& 99.89\%&\textbf{100.00\%} &\textbf{100.00\%} &-&-&-&$O(N^2)$&Node-sequential\\\hline        
        MolGAN~\cite{de2018molgan}&-&-&-&10.40\%&94.10\%&98.10\%&$O(N^2)$&One-shot\\\hline
        MPGVAE\cite{flam2020graph}&-&-&-&68.00\%&54.00\%&91.00\%&$O(N^2)$&One-shot\\\hline
        SCAT\cite{zou2019encoding}&-&-&-&98.30\%&92.00\%&47.40\%&$O(N^2)$&One-shot\\\hline\bottomrule
    \end{tabular}
    \label{tab:comparison_molecule}
\end{table*}\normalsize  

\subsection{Datasets}
The existing benchmark datasets that are typically used in this domain can be categorized into synthetic datasets and real-world datasets. 
We have collected and published all the datasets via this link: \url{https://github.com/xguo7/Dataset-for-Deep-Graph-Generation}.

\subsubsection{Synthetic Datasets}
Followings show the synthetic graph datasets that are typically used in the existing methods. 

\textbf{Community}. It contains 500 two-community graphs with $60\leq|\mathcal{V}|\leq160$. $\mathcal{V}$ denotes the node set of a graph. Each community is generated by the Erdős–Rényi model (E-R)~\cite{erdos1959random} with $n = |V |/2$ nodes and the link probability of $p = 0.3$. 

\textbf{Grid}. It contains 100 standard 2D grid graphs with $100\leq|\mathcal{V}|\leq 400$ and 100 standard large 2D grid graphs with $1296\leq|\mathcal{V}|\leq 2025$.

\textbf{Ego}. It contains 757 3-hop ego networks extracted from the Citeseer network~\cite{sen2008collective} with $50 \leq |\mathcal{V}|\leq399$. Nodes represent documents and edges represent citation relationships.

\textbf{B-A}. 500 graphs with $100\leq|\mathcal{V}|\leq200$ that are generated
using the Barabasi-Albert model. During generation, each
node is connected to 4 existing nodes.

\textbf{Cycles}. A synthetic dataset of graphs with cyclically connected nodes. Each graph is a path with its two end-nodes connected. 500 graphs are generated with size of $10\leq|\mathcal{V}|\leq 100$. 

\textbf{Trees}. A synthetic dataset of 500 trees ($10\leq|\mathcal{V}|\leq 100$) with power law degree distributions. To generate a tree, a trial power law degree sequence is chosen and then elements are swapped with new elements from a powerlaw distribution until the sequence makes a tree.

\textbf{Ladder}. A synsthtic dataset of ladder graphs with $10\leq|\mathcal{V}|\leq 100$, resulting in a total size of 180
graphs. This is two paths of $|\mathcal{V}|/2$ nodes, with each pair connected by a single edge.

\subsubsection{Real-world Datasets}
Followings shows the real-world dataset that are typically used in the existing methods.

\textbf{QM9}~\cite{ramakrishnan2014quantum}. It is an enumeration of around 134k stable organic molecules with up to 9 heavy atoms (carbon, oxygen, nitrogen and fluorine). As no filtering is applied, the molecules in this dataset only reflect basic structural constraints.

\textbf{ZINC}~\cite{irwin2012zinc}. This dataset is a curated set of 250k commercially available drug-like chemical compounds. On average, these molecules are bigger (about 23 heavy atoms) and structurally more
complex than the molecules in \textit{QM9}. 

\textbf{CEPDB}~\cite{hachmann2011harvard}. This dataset consists of organic molecules with an emphasis on photo-voltaic applications. The contained molecules have 28 heavy atoms on average and contain six to
seven rings each.

\textbf{Protein}~\cite{dobson2003distinguishing}. This dataset contains 918 protein graphs with
$100\leq|\mathcal{V}|\leq500$. Each protein is represented by a graph, where nodes are amino acids and two nodes are connected if they are less than 6 Angstroms apart.

\textbf{Enzymes}~\cite{schomburg2004brenda}. This dataset contains protein tertiary structures representing 600 enzymes. Nodes in a graph (protein) represent secondary structure elements, and two nodes are connected if the corresponding elements are interacting. The node labels indicate the type of secondary structure, which is either helices, turns, or sheets.

\textbf{Citation graphs}~\cite{sen2008collective}: Cora and Citeseer are citation networks; nodes correspond to publications and an edge represents one
paper citing the other. Node labels represent the publication area. The Cora dataset contains 2708 nodes, 5429 edges, 7 classes and 1433 features per node. The Citeseer
dataset contains 3327 nodes, 4732 edges, 6 classes and 3703 features per node

\subsection{Results}
To compare the different deep generative models on graphs, we summarize their published experimental evaluation results on two main graph generation tasks. One category is domain-agnostic where the general graph generation tasks are conducted without incorporating the domain knowledge to guarantee the special properties of the graphs, such as synthetic graph generation. The other category is domain-aware, such as molecule structure generation, which requires to consider the validness and properties of the generated nodes, edges, and the whole graphs.

\subsubsection{Results on Domain-agnostic Graph Generation}
The aforementioned benchmark datasets are widely used to evaluate and compare the existing methods.
10 state-of-the-art methods have been compared on three highest-frequently used datasets in Table~\ref{tab:comparison_general}. The 10 methods cover both one-shot and sequential generating styles. The three highest-frequently used datasets are \textit{Community}, \textit{Ego}, and \textit{Protein}\footnote{We categorize it into the domain-agnostic task since it is commonly used without considering the protein properties of the graphs.}. In order to evaluate the quality of the generated graphs, the Maximum Mean Discrepancy (MMD)~\cite{you2018graphrnn} metric is utilized to measure the distance between the learn graph distribution and the real graph distribution, regarding node degree, clustering coefficient, as well as average orbit counts statistics, as described in Section 5. In addition, we also compare the time complexity of these models to reflect their efficiency. In the following, we analyze the experimental results by focusing on
a few issues illustrated as below.\\

\noindent\textbf{Comparison between sequential-based and one-shot-based graph generation for domain-agnostic graphs.} Based on experimental results, we observed that the sequential-based methods, especially RNN-based models, deliver a better performance than the one-shot models for many benchmark datasets. As shown in Table~\ref{tab:comparison_general}, GraphRNN and GraphVRNN achieve the best performance on the \textit{Community} and \textit{Ego} datasets, with GraphVRNN returning to the lowest MMD of 0.025, on average, for the \textit{Community} dataset and with the MMD score for the other methods averaging 0.246. For the \textit{Ego} dataset, GraphRNN delivered the lowest average MMD of 0.017, while the MMD scores for other methods is around 0.060. This may be because sequential-based generation is better at modeling the complex dependency among the nodes and edges in a graph as the conditional distribution of each node or edge is modeled given the partially generated graph.\\

\noindent\textbf{Influence of the attention mechanism on graph generation.} The comparison results presented above suggest that the attention mechanism supports better learning for the graph generation process. As shown in Table~\ref{tab:comparison_general}, among the sequential-based graph generation methods, for the \textit{Protein} dataset the attention-based recurrent neural network GRAN and GRAN-I perform turned in the best performance, with the smallest average MMD scores of 0.063 and 0.046, respectively. This is because the attention mechanism helps distinguish multiple newly added nodes and learns different attention weights for different types of edges during the generation process, thus delivering a more powerful learning capability.\\

\noindent\textbf{Experimental comparisons of complexity.}
Based on the complexity results shown in Table~\ref{tab:comparison_general}, most of the graph generation methods have a complexity of $O(N^2)$. In sequential-based graph generation methods, it is possible to improve the scalability of the generation model from $O(N^2)$ to $O(N\cdot|\mathcal{E}|)$ by implementing an permutation invariant strategy, such as GRAN~\cite{liao2019efficient}. This is because the complexity challenge of the graph generation model arises primarily in the node permutation step when calculating the loss function for optimization. In the case of one-shot generation, since each graph is represented as its adjacent matrix, which limits the complexity of $O(N^2)$. 

\subsubsection{Results on Domain-aware Graph Generation}
Among the domain-aware graph generation tasks, molecule generation is the most popularly explored problem that is handled by a large number of works.
13 domain-specific models that have published molecule generation results on two highest-frequently-used benchmark datasets (i.e., \textbf{QM9} and \textbf{ZINC}) are compared, as shown in Table~\ref{tab:comparison_molecule}. Different from domain-agnostic graph generation, the domain-aware graph generation task values most on the validness of the generated graphs as well as their diversity and novelty regarding the requirement of novel structure design. Thus, three metrics, namely, uniqueness, novelty and validness, are utilized. In the following, we also analyze the experimental results by focusing on a few issues illustrated as below.\\

\noindent\textbf{Influence of domain-specific knowledge on molecule generation.} Table~\ref{tab:comparison_molecule} shows the experimental results for both general graph generation techniques (e.g., GraphVAE, GRF, GraphNVP) and domain-specific graph generation techniques (e.g., JT-VAE and GCPN) when dealing with the molecule generation/optimization problem. Based on these results, incorporating domain-specific knowledge in the form of either learning rewards or regularization can help to generate more valid and unique graphs. For example, JT-VAE and GCPN show better performance than the other methods on the ZINC dataset, especially in terms of Uniqueness and Validness. Specifically, for the ZINC dataset, JT-VAE is the only method that achieve 100\% uniqueness, delivering a performance that is about 26.41\% higher than that of the other methods on average. This is because the inclusion of domain-specific knowledge allows the direct optimization of application-specific objectives, while still ensuring that the generated molecules are realistic and satisfy chemical rules.\\

\noindent\textbf{Experimental comparison of complexity on molecule generation.}
As shown in Table~\ref{tab:comparison_molecule}, the motif-sequential based graph generation methods delivered the most efficient generation process with the lowest complexity. 
For example, JT-VAE and GCPN achieved higher scalability (i.e., $O(C)$) than other models by utilizing a motif-based sequential generating technique. This is because the number of generation iterations are reduced considerably by decomposing all of the nodes into several motif groups, thus reducing the complexity to $O(C)$ where $C$ refers to the number of motifs.\\

\noindent\textbf{Experimental comparison of sequential-based and one-shot graph generation for molecule generation.}
As shown in Table~\ref{tab:comparison_molecule}, methods based on node-sequential and motif-sequential generating techniques, such as CGVAE and MolecularRNN, are also powerful ways to generate molecular structures with high uniqueness, novelty and validity compared to the one-shot generation methods. For example, for the ZINC dataset, the average unique and validity achieved by the sequential-based methods are 81.99\% and 86.16\%, respectively, which are 12.34\% and 26.84\% higher than those of one-shot based methods. Sequential generating techniques, especially motif-based sequential techniques, are more effective in molecule generation tasks for two main reasons: (1) sequential-based methods are capable of modeling the distribution of graph size (i.e., the number of nodes), which varies naturally; and (2) generating a graph based on given motifs (i.e., at a coarse-grained level) decreases the risk of obtaining an invalid results compared to generating a graph based on nodes and edges (i.e., at a fine-grained level).

\end{document}